\newif\ifsingle
\singletrue 

\newif\ifFullVersion

\ifsingle
\documentclass[11pt,draftcls,onecolumn,journal]{IEEEtran}

\else		
\documentclass[10pt,final, twocolumn]{IEEEtran}
\fi

\usepackage{bm}
\usepackage{balance}
\usepackage{times}
\usepackage{amsmath,dsfont}
\usepackage{amssymb}
\usepackage{epsfig,verbatim} 
\usepackage{setspace}
\usepackage{color}
\usepackage{cite}
\usepackage{epstopdf}
\usepackage{subcaption}
\usepackage{graphics}
\usepackage{accents}
\usepackage{acronym}
\usepackage[bookmarks,colorlinks]{hyperref}
\usepackage{ulem}
\usepackage{mathtools}
\usepackage{algorithm}
\usepackage{algorithmic}
\usepackage{enumitem}
\usepackage[most]{tcolorbox}
\usepackage{cuted}
\tcbuselibrary{skins}
\usepackage{multicol,lipsum}
\usepackage{wrapfig}
\usepackage{multirow,colortbl,booktabs}
\usepackage{xcolor-material}
\usepackage{tikz}
\usetikzlibrary{fit,positioning,calc,patterns,arrows,shapes.symbols,decorations.shapes,shapes.geometric}
\tikzset{main/.style = {rounded corners,draw,thick,node font={\bf\color{white}},minimum height=10mm,minimum width=40mm}}

\definecolor{NewColor}{rgb}{0,0,0}
\newcommand{\myVec}[1]{{\boldsymbol{#1}}}
\newcommand{\myMat}[1]{{\boldsymbol{#1}}}
\newcommand{\mySet}[1]{\mathcal{#1}}
\newcommand{\myFunc}[1]{\mathcal{#1}}

\newcommand{\dnnParam}{\myVec{\theta}}
\newcommand{\dnnFunc}{\myFunc{F}_{\dnnParam}}

\newcommand{\Input}{\myVec{x}}
\newcommand{\InputSpace}{\mySet{X}}

\newcommand{\Label}{\myVec{s}}
\newcommand{\LabelSpace}{\mySet{S}}

\newcommand{\ObjParam}{\myVec{\theta}^{\rm o}}
\newcommand{\HypParam}{\myVec{\theta}^{\rm h}}

\newcommand{\csMatrix}{\myMat{H}}
\newcommand{\RPCAMat}{\myMat{X}}
\newcommand{\RPCARank}{\myMat{V}}
\newcommand{\RPCASparse}{\myMat{Y}}

\newcommand{\pdf}{p}

\definecolor{mypurple}{rgb}{0.910, 0.910, 0.969}
\definecolor{myblue}{rgb}{0.122, 0.435, 0.698}

\definecolor{myGreen}{RGB}{112, 173, 71}
\definecolor{myBlue}{RGB}{36, 124, 185}
 
\acrodef{rpca}[RPCA]{Robust Principal Component Analysis}
\acrodef{kf}[KF]{Kalman filter} 
\acrodef{pf}[PF]{particle filter} 
\acrodef{mimo}[MIMO]{multiple-input multiple-output} 
\acrodef{kg}[KG]{Kalman gain}
\acrodef{lasso}[LASSO]{least absolute shrinkage and selection operator} 
\acrodef{ekf}[EKF]{extended \ac{kf}}
\acrodef{ukf}[UKF]{unscented \ac{kf}}
\acrodef{ai}[AI]{artificial intelligence} 
\acrodef{dnn}[DNN]{deep neural network} 
\acrodef{cnn}[CNN]{convolutional neural network} 
\acrodef{nn}[NN]{neural network}  
\acrodef{gnn}[GNN]{graph neural network} 
\acrodef{rnn}[RNN]{recurrent neural network} 
\acrodef{fc}[FC]{fully connected} 
\acrodef{mse}[MSE]{mean-squared error}
\acrodef{snr}[SNR]{signal-to-noise ratio}
\acrodef{ss}[SS]{state-space}
\acrodef{ml}[ML]{machine learning}
\acrodef{gru}[GRU]{gated recurrent unit} 
\acrodef{lstm}[LSTM]{long short-term memory} 
\acrodef{ssm}[SSM]{state-space model} 
\acrodef{ista}[ISTA]{iterative soft-thresholding algorithm} 
\acrodef{ml}[ML]{machine learning}
\acrodef{rpca}[RPCA]{robust principal component analysis}

\IEEEoverridecommandlockouts 

\ifsingle

\newcommand{\boxWidth}{\linewidth}
\else

\newcommand{\boxWidth}{2\linewidth}
\fi

\usepackage[all=normal,paragraphs=tight,floats=normal,mathspacing=normal,wordspacing=tight,charwidths=tight,mathdisplays=normal,leading=normal]{savetrees}

\title{Deep Unfolding: Recent Developments,\\ Theory, and Design Guidelines}
\author{
{
    Nir Shlezinger,~\IEEEmembership{Senior Member,~IEEE},
    Santiago Segarra,~\IEEEmembership{Senior Member,~IEEE}, 
    Yi Zhang,~\IEEEmembership{Member,~IEEE}, 
    Dvir Avrahami,~\IEEEmembership{{Student Member},~IEEE},
    Zohar Davidov,~\IEEEmembership{{Student Member},~IEEE}, 
    Tirza Routtenberg,~\IEEEmembership{{Senior Member},~IEEE},
    and Yonina C. Eldar,~\IEEEmembership{Fellow,~IEEE}
}
\thanks{
		 N. Shlezinger, T. Routtenberg, D. Avrahami, and Z. Davidov are with the School of ECE, Ben-Gurion University of the Negev, Be'er-Sheva, Israel (e-mail: \{nirshl; tirzar\}@bgu.ac.il; \{dviravra; zohardav\}@post.bgu.ac.il).
         S. Segarra is with the Department of Electrical and Computer Engineering, Rice University, TX, USA (e-mail: segarra@rice.edu).
		Y. Zhang and Y. C. Eldar are with the Math and CS Faculty, Weizmann Institute of Science, Rehovot, Israel (e-mail:  \{yi.zhang; yonina.eldar\}@weizmann.ac.il).  Y. C. Eldar is also with the Department of Electrical and Computer Engineering, Northeastern University, MA, USA.
This work has been partially supported by  the European Research Council (ERC) under the ERC starting grant nr. 101163973 (FLAIR).}
 \vspace{-0.5cm}
} 

\begin{document}

\maketitle
\pagestyle{plain}
\thispagestyle{plain}

	\vspace{-0.75cm}

\begin{abstract}
Optimization methods play a central role in signal processing, serving as the mathematical foundation for inference, estimation, and control. While classical iterative optimization algorithms provide interpretability and theoretical guarantees, they often rely on surrogate objectives, require careful hyperparameter tuning, and exhibit substantial computational latency. Conversely, \ac{ml} offers powerful data-driven modeling capabilities but lacks the structure, transparency, and efficiency needed for optimization-driven inference. 
{\em Deep unfolding} has recently emerged as a compelling framework that bridges these two paradigms by systematically transforming iterative optimization algorithms into structured, trainable \ac{ml} architectures. 
This article provides a tutorial-style overview of deep unfolding, presenting a unified perspective of methodologies for converting optimization solvers into \ac{ml} models and highlighting their conceptual, theoretical, and practical implications. 
We review the foundations of optimization for inference and for learning, introduce four representative design paradigms for deep unfolding, and discuss the distinctive training schemes that arise from their iterative nature. 
\textcolor{NewColor}{
To concretely illustrate the discussed principles, we use canonical examples of deep unfolding methodologies for representative problems in sparse recovery and communication optimization, while the presented methodology extends well beyond these specific settings.}
Furthermore, we survey recent theoretical advances that establish convergence and generalization guarantees for unfolded optimizers, and provide comparative qualitative and empirical studies illustrating their relative trade-offs in complexity, interpretability, and robustness.  
\end{abstract}

\acresetall

\section{Introduction}
\label{sec:introduction}  

Optimization methods are fundamental to signal processing, serving as the backbone for a wide range of tasks ranging from  signal recovery and parameter estimation to resource allocation and control~\cite{luo2006introduction}. 
Classical signal processing solutions often rely on optimization-based inference rules, where decisions are obtained by minimizing a cost function that mathematically encapsulates the desired performance criterion or estimation goal. 
This optimization-centric formulation enables practitioners to design decision mappings that are grounded in well-understood mathematical principles and to rigorously analyze their behavior under different modeling assumptions. 
The ability to represent inference as the solution to an optimization problem has therefore become one of the cornerstones of modern signal processing design methodologies, underlying diverse applications in communications, imaging, sensing, and control.

A key enabler of optimization-based inference in practice is the use of {\em iterative solvers}. 
These algorithms, which encompass a broad family of methods for tackling complex optimization problems~\cite{boyd2004convex}, are widely employed due to their flexibility and their suitability for high-dimensional and constrained problems that are analytically intractable. 
However, using iterative optimization for inference introduces several limitations. 
First, the objectives they minimize are often {\em surrogate} or simplified mathematical formulations that only approximate the true performance criterion, resulting in a potential mismatch between the modeled problem and the actual task~\cite{shlezinger2022model}. 
Second, iterative solvers rely on numerous hyperparameters that must be tediously tuned to ensure convergence and desirable performance. 
Finally, reaching convergence may require many iterations, leading to substantial latency and computational overhead, particularly in real-time and large-scale signal processing applications.

In parallel, the rise of \ac{ml}, and particularly deep learning, as the enabler technology for modern \ac{ai},  has revolutionized data-driven modeling across a variety of domains. 
By training highly parameterized architectures end-to-end on large datasets, deep learning enables the construction of powerful mappings that do not require explicit modeling of the underlying physical or statistical process~\cite{dahrouj2021overview}. 
This capability has motivated growing interest in applying \acp{dnn} for tasks that have traditionally been solved via optimization in signal processing~\cite{zappone2019wireless}.
However, while deep learning offers unprecedented empirical success in various domains, most notably in natural language processing and computer vision, its conventional black-box nature  introduces several critical drawbacks when importing such tools for tackling optimization problems in signal processing. 
\acp{dnn} typically lack interpretability, making their internal operations and predictions difficult to explain, analyze, or verify. 
Their reliance on massive labeled datasets limits their applicability in many signal processing settings, where data collection is costly or domain-specific. 
Furthermore, both training and inference often involve heavy computational demands, hindering deployment under tight latency, power, or hardware constraints. 
As such, while deep learning provides an appealing data-driven alternative, its lack of structure and interpretability can make it poorly aligned with the domain knowledge and efficiency requirements that characterize optimization-driven signal processing systems.

 Deep unfolding has emerged in recent years as a compelling methodology for bridging the gap between model-based optimization and deep learning~\cite{hershey2014deep}. 
By systematically transforming iterative optimization algorithms into structured \ac{ml} architectures, deep unfolding combines the interpretability and domain-specific rigor of classical solvers with the abstractness, flexibility, and learning capacity of data-driven learning~\cite{monga2021algorithm}. 
The idea of unfolding dates back to the seminal work of Gregor and LeCun in 2010~\cite{gregor2010learning}, which predated the deep learning revolution sparked by AlexNet~\cite{krizhevsky2012imagenet}. 
However, it is only in the past few years that the methodology has gained substantial traction in the signal processing community, where iterative optimization plays a central role in inference and estimation. 
This surge of interest reflects a growing recognition that deep unfolding offers a principled path for designing {\em learned optimizers}, i.e., systems that not only learn to solve optimization problems efficiently but also adapt their solvers themselves to the underlying task and data~\cite{shlezinger2023model}.

In particular, the growing popularity of deep unfolding has given rise to a rich landscape of methodologies and insights. 
Numerous application-specific realizations have emerged, each tailored to distinct tasks and embodying unique trade-offs between interpretability, flexibility, and complexity. 
Even when starting from the same underlying optimization solver, different unfolding formulations can yield remarkably different \ac{ml} architectures: some focus on learning solver hyperparameters~\cite{khani2020adaptive}, others introduce learned correction terms~\cite{chowdhury2021unfolding}, or augment each iteration with dedicated \acp{dnn}~\cite{samuel2019learning,shlezinger2020deepsic}. 
Beyond these architectural variations, recent research has revealed a growing number of properties stemming from viewing optimization as a learnable process, including inherent scalability~\cite{lavi2023learn}, suitability for distributed and online operation~\cite{noah2024distributed,raviv2023online,saravanos2024deep}, the ability to compensate for computational approximations~\cite{avrahami2025deep}, and even connections to robustness and adversarial sensitivity~\cite{sofer2025unveiling}. 
In parallel, theoretical understanding of deep unfolding has  also advanced considerably, yielding new insights into its convergence~\cite{hadou2024robust}, generalization behavior~\cite{shah2024optimization}, and its ability to interpolate between classical optimization and end-to-end data-driven learning~\cite{scarlett2022theoretical,nareddy2025some}. 
This rapidly evolving body of work motivates the need for a unified treatment that organizes the various forms of deep unfolding methodologies, clarifies their conceptual and theoretical underpinnings, and provides systematic guidance for their design and application in signal processing.

In this article, we provide a tutorial-style overview of {\em{deep unfolding}}, combining a systematic taxonomy, a discussion of recent developments, and insights into the conceptual and theoretical advances that have shaped the field. 
Our goal is to provide an intuitive yet rigorous understanding of the capabilities and possibilities that arise when converting iterative optimization algorithms into \ac{ml} architectures.  
We emphasize both the potential benefits of such hybrid designs, in terms of key measures such as interpretability, efficiency, and adaptability, as well as common challenges arising from their usage for inference in signal processing tasks. 
\textcolor{NewColor}{We thus provide a tutorial-style treatment of deep unfolding that synthesizes recent methodological, theoretical, and practical developments into a unified and application-oriented framework, rather than a comprehensive survey of all related model-based learning paradigms.}
To that end, we begin by establishing the necessary preliminaries in optimization, focusing on the mathematical aspects that are most relevant for understanding the various forms of deep unfolding, and particularly on the formulations of optimization for inference and optimization for learning that serve as its conceptual foundation.

The main body of the article is dedicated to a systematic presentation of deep unfolding methodologies and their categorization according to how iterative optimizers are transformed into trainable \ac{ml} architectures. 
We outline four design approaches that vary in specificity and level of abstraction, each differing in how the underlying optimization process is parameterized, learned, or augmented by neural components. 
For each approach, we present its mathematical formulation, a representative example, and a discussion of the key benefits. 
In addition, we describe the distinct training paradigms that naturally emerge from the structure of deep unfolded architectures, highlighting how their optimization-based origins support flexible learning strategies (both supervised and unsupervised) as well as unique loss formulations that exploit the interpretability of intermediate iterations.

Beyond their design and training, we examine deep unfolding methodologies through three complementary lenses. 
First, we review recent {\em theoretical advances} that provide rigorous analysis and performance guarantees for deep unfolded optimizers, and connect these insights to the methodological frameworks presented earlier. 
Second, we provide a {\em qualitative comparative study}, discussing the relationships among the different unfolding paradigms and their relative gains in terms of complexity, interpretability, robustness to model mismatches, flexibility, and scalability. 
Third, we complement these perspectives with an {\em empirical study} that enables a didactic, side-by-side comparison of the various deep unfolding approaches. 
Specifically, we consider a representative example of an inference task guided by \ac{rpca} optimization~\cite{candes2011robust}.
For this example, we evaluate deep unfolded designs corresponding to each of the considered methodologies in a controlled setting that allows us to isolate and assess the impact of each unfolding strategy on performance metrics such as reconstruction accuracy per iteration and computational complexity. 
The article concludes with a discussion of open challenges and promising directions for future research in the design and application of deep unfolded optimizers.

{\bf Notations:} We use boldface lowercase for vectors, e.g., $\myVec{x}$, and boldface uppercase letters, e.g., $\myMat{X}$, for matrices.  
We use $\mathcal{N}(\myVec{\mu},\myMat{\Sigma})$ for the multivariate Gaussian distribution with mean $\myVec{\mu}$ and covariance $\myMat{\Sigma}$, while $\pdf(\cdot)$ is a probability density function, and $\Vert \myVec{x} \Vert_p$ is the $\ell_p$ norm of $\myVec{x}$. The operator $(\cdot)^\top$ denotes the transpose, while ${\mathbb{R}}$ is the set of real numbers. 

\section{Optimization Fundamentals} \label{sec:fund}

In this section, we present the necessary preliminaries constituting the basis of our article.   The role of this section is to establish the considered framework, while pinpointing the aspects that serve as the basis for the different forms of deep unfolding detailed in the subsequent sections.

\subsection{Optimization Problems}
\label{ssec:fund_OptProb}

{\bf Definition}:
Optimization problems take the generic form  
\begin{align*}
&\mathop{\rm minimize}\limits_{\myVec{s}\in{\mathbb{R}}^n} \quad  \myVec{L}_0(\myVec{s}), \quad
{\rm subject ~to}\quad  \mySet{L}_i(\myVec{s}) \leq 0, \quad i=1,\ldots, m,
\end{align*}
where $\myVec{s}\in{\mathbb{R}}^n$ is the optimization (decision) variable, the function $\mySet{L}_0:{\mathbb{R}}^n\mapsto{\mathbb{R}}$ is the objective (cost) function, and $\mySet{L}_i:{\mathbb{R}}^n\mapsto{\mathbb{R}}$ for $i=1,\ldots, m$ are inequality constraint functions. 
By defining the feasible set $\mySet{S} = \{\myVec{s}: \mySet{L}_i(\myVec{s}) \leq 0, \forall i=1,\ldots, m\}\subset {\mathbb{R}}^n$ (assumed to be nonempty), we can express the global solution to the optimization problem  (assuming a finite solution exists) as 
\begin{equation}
\label{eqn:Opt1}
\myVec{s}^* = \mathop{\arg \min}\limits_{\myVec{s} \in \mySet{S}}\mySet{L}_0(\myVec{s}).
\end{equation}

As the above formulation is extremely generic, it encompasses many important classes of optimization problems. 
One of them is that of {\em linear programming}, for which all the functions $\{\mySet{L}_i(\cdot)\}_{i=0}^m$ are  linear \cite{Luenberger2008}, namely
\begin{equation*}
\mySet{L}_i(\alpha \myVec{s}_1 + \beta \myVec{s}_2) = \alpha \mySet{L}_i( \myVec{s}_1) + \beta \mySet{L}_i (\myVec{s}_2), \qquad \forall \alpha, \beta \in {\mathbb{R}}, \, \myVec{s}_1, \myVec{s}_2 \in {\mathbb{R}}^n.
\end{equation*}
A broader class of problems, which generalizes the case of linear programming, is that of {\em convex optimization}, where for each $i=0,1,\ldots,m$ it holds that
\begin{equation}
\label{eqn:convex}
\mySet{L}_i(\alpha \myVec{s}_1 + (1-\alpha) \myVec{s}_2) \leq \alpha \mySet{L}_i( \myVec{s}_1) + (1-\alpha) \mySet{L}_i (\myVec{s}_2), \qquad \forall \alpha\in [0,1],~ \myVec{s}_1, \myVec{s}_2 \in {\mathbb{R}}^n.
\end{equation}
In particular, if \eqref{eqn:convex} holds for $i=0$, we say that the optimization problem \eqref{eqn:Opt1} has a {\em convex objective}, while when it holds for all $i=1,\ldots, m$ we say that the feasible set $\mySet{S}$ is a {\em convex set}.

{\bf Optimization for Inference}: 
The definition of optimization problems above does not specify the role or interpretation of the objective function. In this article, we focus on optimization problems that correspond to {\em inference} problems or {\em decision rules}. In general, a decision rule $\mySet{F}$ maps the context, denoted $\Input\in\InputSpace$, which is the available observations, into an estimate or decision denoted as $\hat{\Label}\in \LabelSpace$. 	In many signal processing tasks, the inference rule itself can be written as the solution to an optimization problem.

 In such settings, one first defines an objective function $\mySet{L}_{\ObjParam}(\Label;\Input)$, which depends on the available observations (or context) $\Input$ and a set of objective parameters $\ObjParam$. This objective represents a mathematical formulation of the underlying task, and captures both prior understanding of the system and modeling assumptions.  These assumptions may be {\em Bayesian}, in which case $\ObjParam$ includes statistical priors over the unknown quantities, or {\em non-Bayesian (frequentist)}, in which case $\ObjParam$ typically encodes structural constraints, physical parameters, or regularization terms \cite{Kayestimation}.
  In both cases, $\ObjParam$ governs the
 desired trade-offs between fidelity, robustness/regularity, and tractability. 
A representative example appears in the accompanying Box entitled \textit{``Super-Resolution as Sparse Recovery''} on Page~\pageref{Box:SparseRec}, which illustrates how the task of reconstructing high-resolution signals from coarse observations can be cast as an optimization problem with objective parameters $\ObjParam$ that incorporate physical insights and computational feasibility.

 Once the optimization framework is specified, the corresponding inference rule is obtained as the output of an optimization solver that seeks
\begin{equation}
{\Label}^\star = \mathop{\arg\min}\limits_{\Label \in \LabelSpace} \mySet{L}_{\ObjParam}(\Label;\Input),
\label{eqn:OptInference}
\end{equation}
or an approximated solution thereof. This broad formulation highlights the key principle that inference can be expressed as an optimization-driven mapping from data to decisions.

\begin{tcolorbox}[float*=t,
    width=\boxWidth,
	toprule = 0mm,
	bottomrule = 0mm,
	leftrule = 0mm,
	rightrule = 0mm,
	arc = 0mm,
	colframe = myblue,
	colback = mypurple,
	fonttitle = \sffamily\bfseries\large,
	title = Super-Resolution as Sparse Recovery]	
	\label{Box:SparseRec} 
    A representative inference task is super-resolution \cite{EldarKutyniok2012}. 
Here, the goal is to recover a high-resolution image from 
 a distorted low-resolution version. Accordingly,  $\LabelSpace$ and  $\InputSpace$ are the spaces of high-resolution and low-resolution images, respectively. The goal of the decision rule is thus to infer  the high-resolution data from its noisy compressed version $\Input$. 
    
    A popular approach to cast such a setting as an optimization problem is to impose sparsity in some known domain $\myMat{\Psi}$ (e.g., wavelets), such that the high resolution output can be written as $ \myMat{\Psi}\Label$, where $\Label$ is sparse.
    Then, the relationship between the low- and high-resolution images is modeled using a linear transformation $\csMatrix$ (e.g., blur + downsampling). Relaxing the sparsity constraint using an $\ell_1$ regularization leads to the objective
	\begin{equation}
	\label{eqn:InvesObj1}
	\mySet{L}_{ \ObjParam}(\Label; \Input) =   
	\frac{1}{2}\|\myVec{x}-\csMatrix\myMat{\Psi}\Label\|^2_2 +\rho\|\Label\|_1.
	\end{equation}
    Note that the formulation in \eqref{eqn:InvesObj1} is parameterized, with the objective parameters being $\ObjParam = \{\csMatrix,\myMat{\Psi},\rho\}$.
    Here, $\csMatrix$ and $\myMat{\Psi}$ represent the physical and structural models, while $\rho$ balances fidelity and sparsity. 
Solving \eqref{eqn:InvesObj1} yields an estimate ${\Label}^\star=\hat{\Label}$, where $\myMat{\Psi}\hat{\Label}$  serves as the super-resolved reconstruction.

\end{tcolorbox}

{\bf Optimization for Learning}:  
Another common usage of optimization, which is also highly relevant in the context of deep unfolding, arises in {\em learning}. 
Here, unlike in optimization-based inference, the inference rule itself is not defined as the solution to an optimization problem, but rather as a fixed mapping $\dnnFunc:\InputSpace \mapsto \LabelSpace$ parameterized by a set of learnable parameters $\dnnParam$. 
The values of these parameters are determined through a {\em learning phase}, in which a dataset $\mySet{D}$ (consisting of, e.g., representative input–output pairs) is used to construct an {\em empirical risk} function $\mySet{L}_{\mySet{D}}^{\rm ER}(\dnnParam)$ that quantifies how well $\dnnFunc$ aligns with the data. 
The learning process then employs optimization tools to adjust $\dnnParam$ according to
\begin{equation}
\label{eqn:OptLearning}
\dnnParam^* = \arg\min_{\dnnParam\in \Theta} \mySet{L}_{\mySet{D}}^{\rm ER}(\dnnParam),
\end{equation}
so as to yield a mapping $\dnnFunc$ that generalizes well to unseen data, where  $\Theta \subseteq \mathbb{R}^p$ denotes the associated parameter space.

Specifically, the dataset guiding the learning process may be comprised of input-output pairs ($\mySet{D} =\{ \Input_t, \Label_t\}_{t=1}^{|\mySet{D}|}$), as in {\em supervised learning}, or just inputs ($\mySet{D} =\{ \Input_t \}_{t=1}^{|\mySet{D}|}$), as in {\em unsupervised learning}. The empirical risk is typically constructed from $\mySet{D}$ using a sample-wise loss function $l(\cdot)$ (e.g., cross-entropy for classification or $\ell_2$ for regression), along with a regularization term $\Phi(\cdot)$ (e.g., $\ell_p$ regularization) on the trainable parameters. For instance, in a supervised learning setting, the optimization objective in \eqref{eqn:OptLearning} takes the form
\begin{equation}
    \mySet{L}_{\mySet{D}}^{\rm ER}(\dnnParam) = \frac{1}{|\mySet{D}|} \sum_{(\Input_t, \Label_t) \in \mySet{D}} l\left(\dnnFunc(\Input_t), \Label_t\right) + \lambda \cdot \Phi(\dnnParam),
    \label{eqn:SupervisedLoss}
\end{equation}
where $\lambda>0$ controls the regularization strength. 
While the objective parameters $\ObjParam$ are not explicitly written in \eqref{eqn:OptLearning}, empirical risks include such parameters (e.g., $\ObjParam = \lambda$ in \eqref{eqn:SupervisedLoss}).
However, in learning tasks, these parameters are typically introduced to facilitate and stabilize optimization (e.g., regularization coefficients), whereas in optimization for inference $\ObjParam$ often includes parameters stemming from the mathematical description of the task (as in the sparse recovery example).

This formulation highlights the distinction between {\em optimization for inference} and {\em optimization for learning}. 
In optimization for inference, the optimization is executed each time a decision is to be made, directly producing the output $\hat{\Label}$ from the given observations $\Input$. 
In contrast, optimization for learning is typically carried out once during a training stage, producing a parameterized mapping that is subsequently used for inference without solving an optimization problem at test time. 
As will be seen throughout this article, deep unfolding intrinsically combines both aspects: it designs inference rules that are defined via optimization algorithms, yet whose internal parameters are learned from data. 
Still, in the following, we focus our discussion on {\em optimization for inference}, whose fundamental properties and subtleties play a key role in the different forms of deep unfolding.

\subsection{Iterative Optimization}
\label{ssec:fund_OptIter}
The formulation of optimization as inference requires not only crafting an objective $\mySet{L}_{ \ObjParam}(\Label; \Input)$, but also a solver that aims at recovering \eqref{eqn:OptInference} and acts as the decision rule. While the optimization literature includes various forms of solvers, for clarity, we focus our presentation on methods derived from convex problems, as in \eqref{eqn:convex}. 
Convex optimization theory provides useful iterative algorithms for tackling problems of the form \eqref{eqn:Opt1}.  Iterative optimizers typically give rise to additional parameters that affect the speed and convergence rate of the algorithm, but not the actual objective being minimized. We refer to these parameters of the solver as {\em hyperparameters}, and denote them by $\HypParam$. As opposed to the objective parameters $\ObjParam$, they often have no effect on the solution when the algorithm is allowed to run to convergence, and so are of secondary importance.   But when the iterative algorithm is stopped after a predefined number of iterations, it affects the decisions, and therefore also the objective. Due to the surrogate nature of the objective, such stopping does not necessarily degrade the evaluation performance.

For instance, consider an unconstrained optimization in \eqref{eqn:OptInference}, where $\mySet{S}={\mathbb{R}}^n$, 
and the objective $\mySet{L}_{\ObjParam}(\Label;\Input)$ is convex and differentiable w.r.t. $\Label$.
In this setting, one can iteratively update over $\Label^{(k)}$, where $k=0,1,2,\ldots$ is the iteration index, and guarantee that under standard conditions the sequence $\{\Label^{(k)}\}$ 
monotonically does not increase the objective, i.e., $\mySet{L}_{\ObjParam}(\Label^{(k+1)};\Input) \leq \mySet{L}_{\ObjParam}(\Label^{(k)};\Input)$, and converges to a minimizer of the problem, or to the global minimizer $\myVec{s}^*$ when the objective is convex. The family of iterative optimizers that operate this way is referred to as {\em descent methods}.
Descent methods that operate based on first-order derivatives of the objective, i.e., using gradients, are referred to as {\em first-order methods}. Arguably the most common first-order optimizer is {\em gradient descent}, which stems   from the first-order multivariate Taylor series expansion of $\mySet{L}_{\ObjParam}(\Label;\Input)$ around  the current iterate $\Label^{(k)}$,  to  obtain an update equation 
\begin{equation}
\label{eqn:FullGD}
\Label^{(k+1)} \leftarrow \Label^{(k)} - \mu_k \nabla_{\Label} \mySet{L}_{\ObjParam}(\Label^{(k)};\Input),
\end{equation}
 which is applied iteratively for $k=0,1,2,\ldots$. 
Here, $\mu_k$ is referred to as the {\em step size} or the {\em learning rate}, which controls the magnitude of the update at iteration $k$.

In optimization problems where the objective function can be decomposed into the sum of two functions, $g(\cdot)+h(\cdot)$,  with $g(\cdot)$ smooth and $h(\cdot)$ possibly nonsmooth,  a widely-used descent method is based on {\em proximal gradient descent}. 
\textcolor{NewColor}{At each iteration $k$, this method updates $\Label^{(k)}$ by locally approximating the smooth term $g(\cdot)$ with a quadratic upper bound (which can be interpreted as a second-order Taylor approximation with a scaled identity Hessian surrogate), leading to the iterative update rule}
\begin{equation}
\label{eqn:ProxGD}
\Label^{(k+1)} \leftarrow  {\rm prox}_{\mu_k\cdot h}\left( \Label^{(k)} - \mu_k \nabla_{\Label} g(\Label^{(k)})\right).
\end{equation}
In \eqref{eqn:ProxGD} we use ${\rm prox}_{\mu_k\cdot h}$ to represent the proximal mapping, defined as ${\rm prox}_{\phi}(\myVec{y})\triangleq \mathop{\arg \min}_{\myVec{z}}\frac{1}{2}\|\myVec{z}-\myVec{y}\|_2^2 + \phi(\myVec{z})$,  with $\phi(\myVec{z})\equiv \mu_k \cdot h(\myVec{z})$. 
\textcolor{NewColor}{Intuitively, the proximal operator enables the use of gradient-based optimization tools for composite objectives in which one term is smooth and differentiable while the other may be nonsmooth. Rather than taking a gradient step on the nonsmooth term, the update combines a gradient step on the smooth component with a proximal mapping that explicitly accounts for the structure of the nonsmooth regularization.}
In the context of our previous example of sparse recovery, a common special case of proximal gradient descent is recalled in the box entitled {\em  \Acl{ista}} on Page~\pageref{Box:ProxGrad}.

\begin{tcolorbox}[float*=t,
    width=\boxWidth,
	toprule = 0mm,
	bottomrule = 0mm,
	leftrule = 0mm,
	rightrule = 0mm,
	arc = 0mm,
	colframe = myblue,
	colback = mypurple,
	fonttitle = \sffamily\bfseries\large,
	title = \Acf{ista} for sparse super-resolution]	
	\label{Box:ProxGrad}

Recall the $\ell_1$ regularized objective used for describing the super-resolution with a sparse prior in \eqref{eqn:InvesObj1}, where, for brevity, we consider sparsity in the signal domain, i.e., $\myMat{\Psi} =\myMat{I}$. 
In this case, the objective in \eqref{eqn:InvesObj1} takes the form  $g(\cdot)+h(\cdot)$ with $h(\Label) \equiv \rho\|\Label\|_1$ and $g(\Label) \equiv \frac{1}{2}\|\myVec{x}-\csMatrix\Label\|^2$. The corresponding proximal mapping is given by
	\begin{align}
	{\rm prox}_{\mu_k\cdot h}(\myVec{y}) = \mathop{\arg \min}_{\myVec{z}}\frac{1}{2}\|\myVec{z}-\myVec{y}\|_2^2 + \mu_k\rho\|\myVec{z}\|_1  &= \mySet{T}_{\mu_k\rho} (\myVec{y}),
	\end{align}
	where $\mySet{T}_{\mu_k\rho}(\cdot)$ denotes the soft-thresholding operation applied element-wise. 
    Substituting this proximal mapping and the gradient of $g(\cdot)$, given by 
$\nabla_{\myVec{s}} g(\myVec{s}) = \csMatrix^T(\csMatrix\myVec{s} - \myVec{x})$,  into the proximal gradient descent update in \eqref{eqn:ProxGD} yields the well-known {\em \ac{ista}}, whose update equation is given by
	\begin{equation}
	\label{eqn:ISTA}
	\Label^{(k+1)} \leftarrow  \mySet{T}_{\mu_k\rho}\left( \Label^{(k)} + \mu_k \csMatrix^T(\myVec{x}-\csMatrix\Label^{(k)}) \right).
	\end{equation}
    The resulting inference rule has objective parameters $\ObjParam = \{\csMatrix, \rho\}$ and  hyperparameters $\HypParam = \{\mu_k\}$.
\end{tcolorbox}

{\bf Pros and Cons of Iterative Optimization-Based Inference}:
The examples above represent only a small subset of the rich family of iterative convex optimizers, which, on its own, is only a portion of the vast landscape of the optimization literature. Nevertheless, it provides us with the understanding of the existing parameterization of iterative solvers via $\ObjParam$ and $\HypParam$, as well as the motivation for deep unfolding, which arises from the following characterization of the pros and cons of iterative optimization-based inference rules.

Optimization-based inference rules, in which decisions are obtained as the solutions to optimization problems, exhibit several  desirable properties that motivate their widespread use in signal processing and related fields:
\begin{enumerate}[label={P\arabic*}]
\item \label{itm:Optimality} {\em Optimality} - When the considered task can be faithfully formulated as a convex optimization problem, iterative descent methods are guaranteed to converge to its global minimum. In such cases, the resulting inference rule is optimal in the sense that it produces the most desirable decision according to the established formulation of the task.

\item \label{itm:Adaptability} {\em Adaptability} - A major advantage of iterative optimization-based inference is its inherent adaptability. Once an iterative solver is devised, it can typically be applied to different parameterizations of the objective (i.e., varying $\ObjParam$)
without redesigning the algorithmic structure. For example, the same \ac{ista} solver can be used for sparse recovery  under different measurement matrices~$\csMatrix$, where changing~$\csMatrix$ merely modifies the instantiated objective~$\mySet{L}_{\ObjParam}(\Label;\Input)$ while preserving the solver’s operational flow.

\item \label{itm:interpretability} {\em Interpretability} - Iterative optimization-based inference rules are also highly interpretable. Each iteration has a clear mathematical meaning and can be analyzed to reveal how information is refined from one step to the next. This interpretability enables one to explicitly monitor the intermediate features exchanged across iterations, which often correspond to progressively improved estimates of the desired decision. Such transparency allows practitioners to reason about the algorithm’s operation, convergence, and sensitivity to parameters.

\item \label{itm:Scalability} {\em Scalability} - Iterative optimizers are inherently scalable. Once an update rule is established, it can be applied seamlessly to problems of different dimensions, i.e., for varying signal sizes~$n$ and measurement dimensions~$m$. This dimension-agnostic nature facilitates the use of the same optimization-based inference mechanism across a broad range of problem instances without requiring any retraining or structural modification. \textcolor{NewColor}{While scalability is not unique to optimization-based inference (e.g., convolutional or graph neural networks can also operate across varying spatial or graph dimensions), optimization-derived update rules are often inherently dimension-agnostic, allowing the same solver structure to be directly applied to problem instances of different sizes without architectural modification.}
\end{enumerate}

While Properties \ref{itm:Optimality}-\ref{itm:Scalability} make optimization-based inference rules highly appealing, their practical deployment in real-world applications often encounters several challenges. These include:
\begin{enumerate}[label={C\arabic*}]
\item \label{itm:Objective} {\em Surrogate Objectives} - The objective guiding the inference rule is typically a surrogate of the true performance goal. 
For example, while the mathematical formulation of super-resolution as sparse recovery yields a tractable optimization problem, it may not fully capture the complex process of reconstructing high-resolution content from low-resolution, distorted measurements. Consequently, the  solution to the surrogate objective may not directly correspond to optimal task performance in practice.

\item \label{itm:Relaxation} {\em Relaxed Formulations} - 
Even when the surrogate objective adequately models the intended task, it is often further relaxed to ensure tractability. For instance, the use of $\ell_1$-norm regularization to promote sparsity constitutes a convex relaxation of the nonconvex $\ell_0$ constraint that truly represents sparsity. Such relaxations are crucial for computational feasibility but can compromise fidelity to the original problem and increase sensitivity to the choice of the objective parameters~$\ObjParam$.

\item \label{itm:Latency} {\em Latency} -  Achieving convergence often requires a large number of iterations, which can be prohibitive in time-critical applications. Although acceleration schemes, such as momentum-based updates or backtracking line search, can reduce the iteration count, they typically increase the per-iteration computational burden. As a result, the overall latency may remain high, especially in real-time or large-scale systems.

\item \label{itm:Complexity} {\em Computational Complexity} - In addition to latency, the computational complexity of each iteration can itself be substantial. Certain optimization methods involve matrix inversions or other heavy algebraic operations, which may be infeasible on hardware-limited platforms. These computational demands become particularly critical in embedded or distributed signal processing systems, where efficiency, memory, and energy constraints are central considerations.

\end{enumerate}

In summary, while iterative optimization-based inference rules provide optimality, adaptability, interpretability, and scalability, they also face notable limitations related to the surrogate and relaxed nature of their objectives, as well as their computational and latency bottlenecks. These shortcomings motivate the development of hybrid methodologies, especially deep unfolding, that aim to retain the benefits of optimization-based inference while overcoming its practical limitations. 
The above characterization of the desirable properties and the challenges is summarized in Table~\ref{tab:myProsConsSummary}. Specifically, the challenges characterized in \ref{itm:Objective}-\ref{itm:Complexity} 
motivate exploring data-driven approaches for augmenting iterative optimizers and leveraging learning capabilities.

\begin{table}
\centering 
\setlength{\tabcolsep}{2pt} 
\renewcommand{\arraystretch}{1.5}
{\scriptsize
    \begin{tabular}{|p{0.5cm} p{6.5cm}| p{0.5cm} p{6.5cm}|}
    \hline
       \multicolumn{2}{|c|}{{\bf Desirable Properties}} &  \multicolumn{2}{|c|}{{\bf Challenges}}\\
       \hline
       \hline
       \ref{itm:Optimality}  & Recover optimal solution under some conditions 
       & \ref{itm:Objective} &  Objective is often a surrogate of the true task  \\
       \ref{itm:Adaptability} & Adaptable to different objectives 
       & \ref{itm:Relaxation} & Objectives are often relaxed for tractability \\
       \ref{itm:interpretability} & Interpretable operation & 
       \ref{itm:Latency} & High latency due to numerous iterations \\
       \ref{itm:Scalability} & Scalable to different data dimensions &
       \ref{itm:Complexity} & Each iteration can involve complex computations \\ 
       \hline
    \end{tabular}
}
    \caption{Summary of desired properties and challenges of iterative optimizers.}
    \label{tab:myProsConsSummary}
\end{table}






\section{Deep Unfolding} 
\label{sec:Unfolding} 
In this section, we present the core concept of {\em deep unfolding} and the principal design methodologies that arise when iterative optimization algorithms are converted into trainable \ac{ml} architectures. Deep unfolding provides a systematic 
framework that bridges model-based optimization and data-driven learning, retaining the interpretability and theoretical grounding of classical solvers with the flexibility and adaptability of deep learning. The section is organized as follows: we first describe the underlying rationale of deep unfolding, establishing how it unifies the principles of optimization for inference and optimization for learning. Then, we detail four representative paradigms for parameterizing architectures, each accompanied by its mathematical formulation, a concrete example, and a discussion of its main advantages and trade-offs. We conclude with an overview of the various training strategies supported by unfolded optimizers.

\begin{figure}
    \centering
    \includegraphics[width=1.05\linewidth]{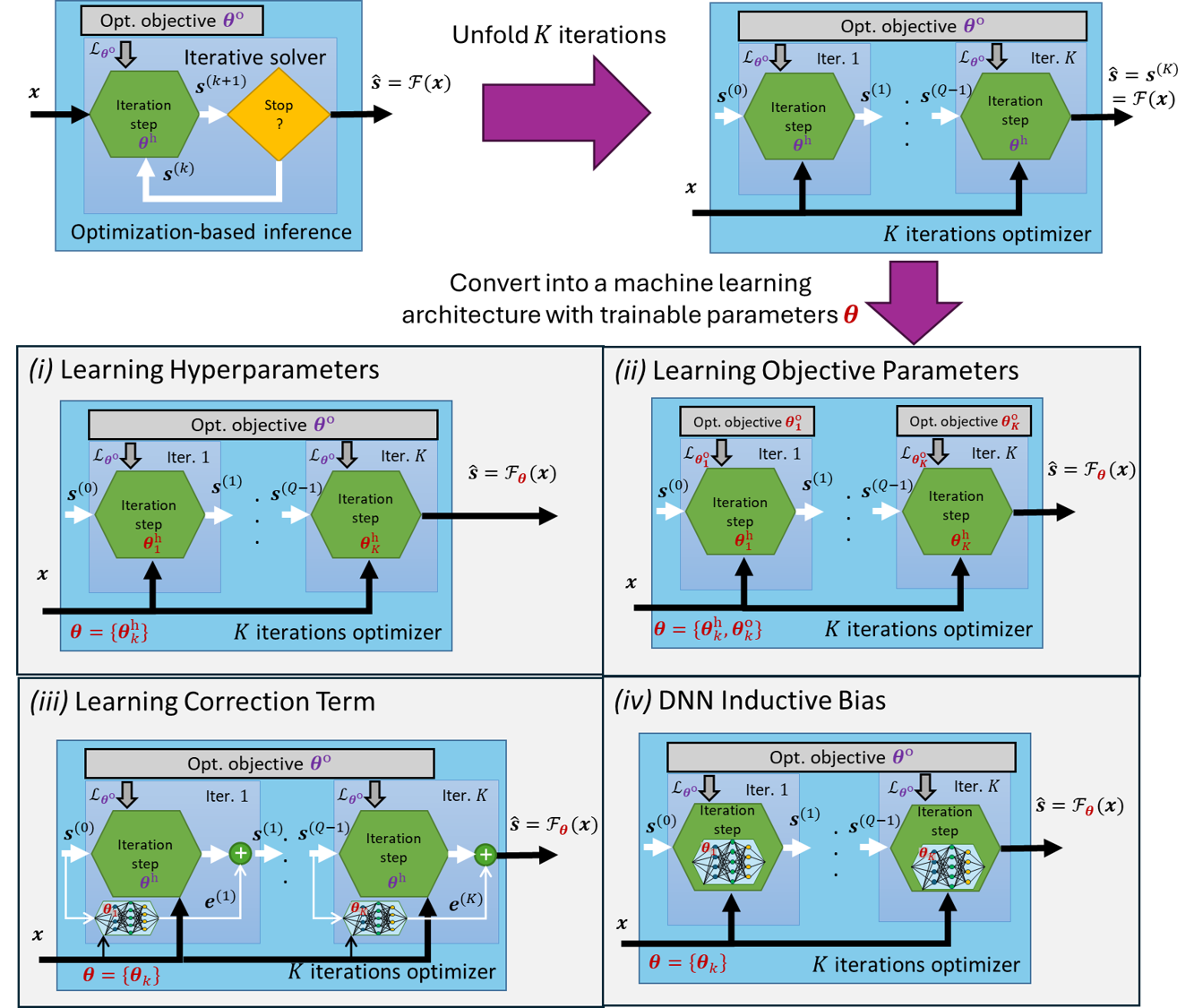}
    \caption{Illustration of considered methodologies for converting an iterative optimization-based inference mapping into an unfolded \ac{ml} architecture. Trainable parameters are marked in \textcolor{red}{red font}. 
    }
    \label{fig:UnfoldArch1}
\end{figure}

\subsection{Rationale}
\label{ssec:DU_Rationale}

The essence of deep unfolding is straightforward: it fuses {\em optimization for inference} and {\em optimization for learning} by transforming an iterative optimization algorithm into a structured, trainable model. As its name suggests, the approach relies on {\em unfolding} the iterations of an optimization algorithm that is originally designed to solve a specific inference task, into a sequential computational graph with a fixed number of iterations. Each iteration is then parameterized and represented as a layer of a \ac{dnn}, where the algorithm’s operations (e.g., gradient steps, thresholds, or filters) are implemented as differentiable modules with learnable parameters.
These parameters can be optimized using data-driven learning techniques.

\textcolor{NewColor}{Deep unfolding is part of a broader family of approaches often referred to as {\em learn-to-optimize}~\cite{chen2022learning,song2024towards}, which combine \ac{ml} with iterative optimization algorithms. Beyond finite-depth unrolling, this family includes implicit-layer approaches such as deep equilibrium models~\cite{bai2019deep}, as well as plug-and-play methods~\cite{ahmad2020plug} and related algorithm-inspired architectures~\cite{shlezinger2023model} that embed learned modules within iterative optimization frameworks. The distinguishing feature of deep unfolding is the explicit truncation of an iterative solver into a fixed-depth architecture trained end-to-end, whereas implicit or plug-and-play paradigms typically preserve the convergence-oriented viewpoint and differentiate through fixed-point conditions or outer optimization loops. In this article, we focus specifically on the deep unfolding paradigm and its unique properties.}

\textcolor{NewColor}{The guiding rationale of deep unfolding} encompasses a remarkably broad family of learned optimizers that differ substantially in their specificity, degree of abstraction, and parameterization. This diversity arises from the dual parameterization inherent to optimization-based inference: the {\em objective parameters}~$\ObjParam$, which characterize the mathematical formulation of the task, and the {\em hyperparameters}~$\HypParam$, which govern the behavior of the iterative solver. Depending on which of these elements are made learnable and how learning is integrated into the iterative process, different forms of deep unfolding emerge, each suited to distinct scenarios and offering unique benefits. \textcolor{NewColor}{The selection of the underlying iterative solver is a design choice on its own. A given optimization problem can often be addressed by multiple algorithms, each inducing a distinct computational structure and parameterization. When unfolded, these structural differences directly affect the resulting architecture, including the number of learnable hyperparameters and the degree of abstractness.  Accordingly, the choice of solver is not merely an optimization detail, but a fundamental architectural decision in deep unfolding design.}

{\textcolor{NewColor}{\textbf{For a given iterative optimizer}}, one can identify four principal paradigms of deep unfolding: 
\begin{itemize}
    \item[$(i)$] {\em Learning Hyperparameters}, which focuses on learning iteration-specific solver parameters (e.g., step sizes, penalties, or momentum coefficients);
    \item[$(ii)$] {\em Learning Objective Parameters}, which extends this approach by jointly learning the parameters defining the optimization objective itself at each iteration;
    \item[$(iii)$] {\em Learning Correction Term}, which employs deep learning tools to inject corrective terms into the iterative updates, improving robustness and flexibility; and 
    \item[$(iv)$] {\em \ac{dnn} Inductive Bias}, which replaces internal computations of the optimization process with compact neural modules, yielding modular architectures that retain the optimization structure while enhancing expressivity, and can operate with reduced complexity and faster inference compared to classical optimizers in complex problems. 
\end{itemize}
An illustration of the different architectural design approaches is depicted in Fig.~\ref{fig:UnfoldArch1}. All methods result in the iterative optimizer with $K$ iterations being cast as an \ac{ml} model. 
In the following subsections, we elaborate on each of these methodologies, providing their mathematical formulations, representative examples, and insights into their practical and theoretical gains, followed by a comparative discussion in Section~\ref{sec:Comparative}. 
\textcolor{NewColor}{While, for clarity of exposition, the illustrative examples in the sequel focus on optimization problems written in regularized (unconstrained) form, the deep unfolding methodologies described in this section apply equally to minimax~\cite{lavi2023learn} or to explicitly constrained formulations (as in \eqref{eqn:Opt1})~\cite{zhang2025feasibility,hadou2025unrolled}, by unfolding the corresponding iterative solvers.}

\subsection{Learning Hyperparameters}
\label{ssec:Unfold_hyp}
 The first approach, often regarded as a {\em shallow} form of deep unfolding, treats the hyperparameters of the iterative solver as trainable parameters. In this setting, the structure of the original optimization algorithm is fully preserved, while its tunable coefficients are optimized from data. This approach thus represents the most direct and interpretable way to convert an iterative optimization procedure into a trainable \ac{ml} model, bridging optimization-based inference and data-driven learning in a principled yet lightweight manner.

\paragraph*{Formulation}
Following the common rationale of deep unfolding,  the iterative optimizer is fixed to operate for a pre-defined number of iterations, denoted by $K$. 
Let $\HypParam_k$ represent the set of hyperparameters used by the solver at iteration $k$. 
The learning-based formulation of the unfolded optimizer considers these $\{\HypParam_k\}_{k=1}^K$ as the trainable parameters of the architecture, i.e., $\dnnParam = \{\HypParam_k\}_{k=1}^K$. 
Each iteration follows the original solver update rule, obtaining $\Label^{(k+1)}$ based on the previous iteration $\Label^{(k)}$ and the input $\Input$ as well as the parameters $ \HypParam_k$ and $\ObjParam$ using the update mapping of the chosen optimizer (e.g., a gradient or proximal step). 
Training is then carried out by applying standard learning procedures over a suitable empirical loss function (as discussed in Subsection~\ref{ssec:Unfolding_Train}), thereby identifying the values of $\{\HypParam_k\}_{k=1}^K$ that yield the best performance on the training data while maintaining a fixed latency (i.e., number of iterations). 
A concrete example of this methodology, based on~\cite{lavi2023learn}, is provided in the box entitled {\em Unfolded Projected Gradient Ascent for Hybrid Beamforming} on Page~\pageref{Box:LearnHypExm}.

Casting hyperparameters as trainable weights provides additional flexibility beyond traditional iterative optimization. 
For instance, instead of relying on manually tuned scalar step sizes, one may adopt multivariate or even structured step-size matrices that vary across iterations, as these parameters are now learned automatically rather than chosen heuristically. 
This parameterization allows the optimizer to adapt its internal dynamics to the data without sacrificing interpretability, complexity, or mathematical grounding.

\begin{tcolorbox}[float*=t,
    width=\boxWidth,
	toprule = 0mm,
	bottomrule = 0mm,
	leftrule = 0mm,
	rightrule = 0mm,
	arc = 0mm,
	colframe = myblue,
	colback = mypurple,
	fonttitle = \sffamily\bfseries\large,
	title = Unfolded projected gradient ascent for hybrid beamforming]	
	\label{Box:LearnHypExm} 
 The hybrid beamforming setting in~\cite{lavi2023learn} considers a base station with $M$ antennas communicating over $B$ frequency bands using $L<M$ RF chains. Transmission precoding thus involves setting the digital precoders $\myMat{W}_{d,b}\in\mathbb{C}^{L\times{N}}$ for each band $b$, where 
$N$ is the number of transmitted data streams, and an analog precoder $\myMat{W}_a\in \mathbb{C}^{M\times{L}}$ whose entries are constrained to have unit magnitude.
For a given channel realization $\Input = \{\myMat{H}_b\}_{b=1}^B$, the precoders $\Label = [\myMat{W}_a,\{\myMat{W}_{d,b}\}]$ are tuned by solving an optimization problem whose objective is to maximize the achievable sum rate,  i.e., 
 \begin{equation}
     \mySet{L}_{\ObjParam}(\Label;\Input) = -R\left(\myMat{W}_a , \{\myMat{W}_{d,b}\}, \{\myMat{H}_b\}\right) \triangleq  \frac{-1}{B} \sum_{b=1}^{B} \log \left|\textbf{I} \!+\! \frac{1}{\sigma^2} \myMat{H}_b\myMat{W}_a\myMat{W}_{d,b} \myMat{W}_{d,b}^H\myMat{W}_a^H\myMat{H}_b^H\right|, \label{eqn:achievable rate}
 \end{equation}
where the objective parameters $\ObjParam$ include the noise level $\sigma^2$.
A candidate iterative optimizer for setting $\Label$ based on \eqref{eqn:achievable rate}  is projected gradient ascent, whose  $k$th iteration updates $\myMat{W}_a$ and  $\myMat{W}_{d,b}$  via
\begin{subequations}
    \label{eqn:PGAsteps}
    \begin{align}
	{\myMat{W}_a^{(k+1)}}
	= \Pi_{a}\Big(\myMat{W}_a^{(k)}
	 &+\mu_a^{(k)}\frac{\partial}{\partial\myMat{W}_a}
	{ R}\Big([\myMat{W}_a^{(k)}, \{\myMat{W}_{d,b}^{(k)}\}], \{\myMat{H}_b\}\Big)\Big),\label{eqn:Wa_step_proj}\\
	  {\myMat{W}}_{d,b}^{(k+1)} 
	= \Pi_{d}\Big(\myMat{W}_{d,b}^{(k)}
  &+ \mu_{d,b}^{(k)}\frac{\partial}{\partial\myMat{W}_{d,b}}
	{ R}\Big([\myMat{W}_a^{(k\!+\!1)} , \{\myMat{W}_{d,b}^{(k)}\}], \{\myMat{H}_b\} \Big)\Big),
\quad b=1,\ldots,B.   	\label{eqn:gr_Wdb_k_step}  
	\end{align}
\end{subequations} 
In \eqref{eqn:PGAsteps}, the hyperparameters $\HypParam_k = [\mu_a^{(k)},\{\mu_{d,b}^{(k)}\}]$ are positive scalar step sizes,
while $\Pi_a(\cdot), \Pi_d(\cdot)$ are projection operators enforcing, respectively, the unit-magnitude of $\myMat{W}_a$ and an overall power constraint on $\{\myMat{W}_{d,b}\}$.

\begin{wrapfigure}{r}{0.7\linewidth} 
    \centering
    \vspace{-1.4cm}
    \includegraphics[width=\linewidth]{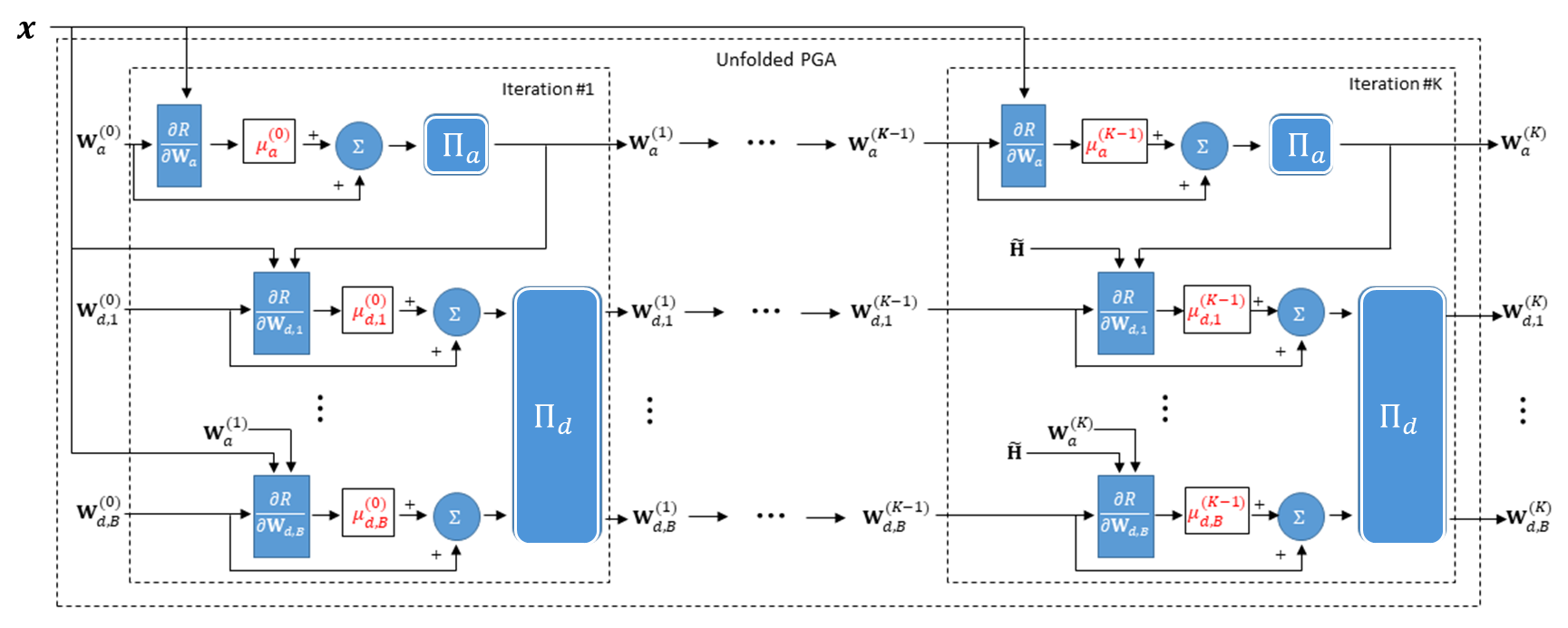} 
    \caption{Unfolded PGA illustration, with $\dnnParam$ marked in \textcolor{red}{red}} 
    \label{fig:UnfoldPGA2}
    \vspace{-0.4cm}
\end{wrapfigure}
As hybrid beamformers must be set rapidly to cope with channel variations, the work \cite{lavi2023learn} proposed to unfold \eqref{eqn:PGAsteps} into $K$ iterations, while using data to learn iteration-specific hyperparameters $\dnnParam = \{\mu_a^{(k)}, \{\mu_{d,b}^{(k)}\}\}$ that maximize the rate \eqref{eqn:achievable rate} within $K$ iterations. The resulting architecture is depicted in Fig.~\ref{fig:UnfoldPGA2}.
\end{tcolorbox}

\paragraph*{Gains}
This form of deep unfolding fully preserves the operation of the original iterative optimizer (thus satisfying  \ref{itm:interpretability}), modifying only its hyperparameters, which are typically manually tuned, while preserving its adaptability to different objective parameters (\ref{itm:Adaptability}). 
Its main motivation lies in enforcing an iterative optimizer to achieve its best possible performance under a fixed and predefined latency constraint (tackling \ref{itm:Latency}), i.e., with a fixed number of iterations~$K$. 
Unlike conventional methods for hyperparameter adaptation, such as backtracking or line search~\cite[Ch. 3]{nocedal1999numerical}, \cite{cavalcanti2025adaptive}, learning the hyperparameters is carried out during an offline training phase, and thus, incurs no additional inference-time complexity.

The ability to optimize an iterative solver for a fixed number of iterations provides benefits beyond latency reduction. 
In distributed optimization, for example, each iteration typically corresponds to a communication round among agents; thus, optimizing an unfolded optimizer with fixed $K$ translates into achieving the best performance for a predetermined number of communication exchanges~\cite{noah2024distributed}. 
Moreover, latency reduction can also be achieved by supporting simplified computations within each iteration (thus tackling \ref{itm:Complexity}). 
It has been shown that using high-dimensional, data-driven hyperparameters can compensate for errors induced by approximate or reduced-complexity operations, enabling efficient inference even under constrained hardware or computational budgets~\cite{avrahami2025deep}.

An additional appealing property of learning hyperparameters, particularly when compared to standard black-box \acp{dnn}, is their inherent {\em scalability} (i.e., \ref{itm:Scalability}). 
Since the structure of the iterative optimizer is independent of the problem dimensions, the learned unfolded model can generalize across different input sizes. 
For example, the learned unfolded hybrid beamformer of \cite{lavi2023learn} trained for a system with a given number of antennas can often be directly applied to systems with different antenna configurations without retraining, thanks to the algorithmic structure inherited from its optimization-based formulation. 
This scalability, coupled with interpretability and low training overhead, makes learning hyperparameters an attractive and practical entry point into the family of deep unfolding methodologies.

\subsection{Learning Objective Parameters}
\label{ssec:Unfolding_obj}
 A more abstract approach for converting iterative optimizers into \ac{ml} models, while still preserving their original operation and without incorporating any artificial neurons, leverages the additional existing parameterization of optimization solvers in their objectives. This methodology provides a higher level of abstraction compared to learning only the solver hyperparameters, while maintaining a clear algorithmic interpretation in which every iteration remains an identifiable and interpretable optimization step. In this sense, this approach extends the concept of data-driven adaptation from tuning the procedural coefficients of the solver to refining the very objectives that guide the optimization, leading to learned optimizers that adapt their internal logic based on data.

\paragraph*{Formulation}
In this methodology, we again consider the original iterative optimizer operating with a fixed number of iterations~$K$. 
However, instead of only learning iteration-dependent hyperparameters, we now let each iteration have its own set of parameters denoted $\dnnParam_k$, which includes both the solver hyperparameters and the parameters of the objective that guides that iteration, i.e., $\dnnParam_k = \{\HypParam_k, \ObjParam_k\}$. 
The resulting deep unfolded model thus learns, from data, both {\em how} to optimize and {\em what} to optimize at each iteration. 
These per-iteration parameters are tuned based on the overall performance of the resulting unfolded optimizer in fitting the training data.
In this manner, each iteration effectively performs a gradual optimization step, possibly guided by a distinct objective function. 
This flexible parameterization allows the unfolded optimizer to mimic an iterative process while operating under a learned, iteration-dependent family of objectives. 
A representative example of this methodology is the original LISTA algorithm, which is described in the box on Page~\pageref{Box:LearnObjExm}.

\begin{tcolorbox}[float*=t,
    width=\boxWidth,
	toprule = 0mm,
	bottomrule = 0mm,
	leftrule = 0mm,
	rightrule = 0mm,
	arc = 0mm,
	colframe = myblue,
	colback = mypurple,
	fonttitle = \sffamily\bfseries\large,
	title = LISTA]	
	\label{Box:LearnObjExm} 
 	Consider again the \ac{ista} optimizer, which solves the convex  LASSO problem in \eqref{eqn:InvesObj1} 
	via the iterative update equations in \eqref{eqn:ISTA}. 	
	The  learned \ac{ista} (LISTA) \ac{ml} architecture, considered the origin of deep unfolding  \cite{gregor2010learning}, unfolds \ac{ista} by fixing $K$ iterations and replacing the update step  with
\begin{wrapfigure}{r}{0.55\linewidth} 
    \centering
    \vspace{-0.3cm}
    \includegraphics[width=\linewidth]{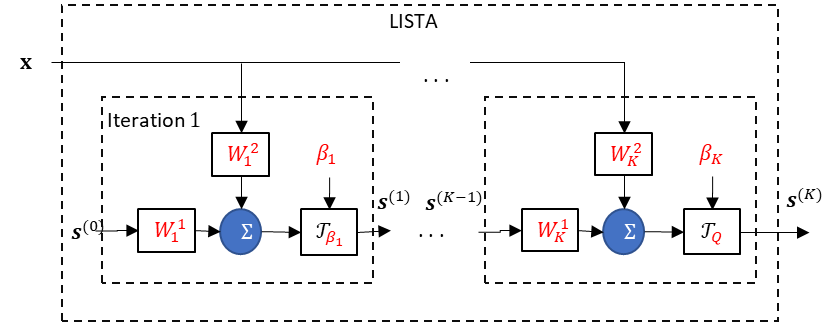} 
    \caption{LISTA illustration, with $\dnnParam$ marked in \textcolor{red}{red} } 
    \label{fig:LISTA1}
\end{wrapfigure}
	\begin{equation}
	\label{eqn:LISTA}
	\Label^{(k+1)} = \mySet{T}_{\beta_k}\left( \myMat{W}_k^1\myVec{x} +  \myMat{W}_k^2  \Label^{(k)} \right).
	\end{equation} 
    The trainable parameters $\myVec{\theta}=\big[\{\myMat{W}_k^1, \myMat{W}_k^2, \beta_k\}_{k=1}^K\big]$ encompass both the solver hyperparameters of \ac{ista} (e.g., $\mu_k$ in \eqref{eqn:InvesObj1}), as well as its objective parameters (i.e., $\csMatrix$ and $\rho$ in \eqref{eqn:ISTA}). The architecture is depicted in Fig.~\ref{fig:LISTA1}. 
	Note that for $\myMat{W}_k^1 = \mu\csMatrix^T$, $\myMat{W}_k^2 = \myMat{I} -\mu\csMatrix^T\csMatrix$,  and $\beta_k=\mu \rho$, \eqref{eqn:LISTA} coincides with the model-based \ac{ista}. 
\end{tcolorbox}

\paragraph*{Gains}
The formulation above reveals two primary gains of this methodology. 
First, as in the case of learning iteration-dependent hyperparameters, this approach facilitates operation with a fixed and pre-defined number of iterations, thereby optimizing the optimizer within a desired latency constraint (\ref{itm:Latency}). 
However, by allowing each iteration to be guided by its own learned objective, we gain additional abstractness and flexibility: the final output after a small number of iterations can perform well with respect to the original optimization goal, even if the intermediate objectives differ. 
This notion of iteration-wise objective adaptation provides a principled way to compress long iterative processes into short and trainable ones without losing interpretability.

\textcolor{NewColor}{
The second key advantage stems from the flexibility to adapt surrogate  (\ref{itm:Objective})  or relaxed objectives (\ref{itm:Relaxation}) to the data. While learning objective parameters does not eliminate the surrogate modeling assumption, it enables iteration-dependent adaptation of the objective formulation. For example, objective parameters, such as regularization coefficients in sparse recovery (e.g., the $\ell_1$ coefficient $\rho$ in ISTA), typically require manual tuning to achieve satisfactory results. In LISTA-type architectures, different layers effectively employ different regularization strengths, allowing the unfolded optimizer to adjust the surrogate trade-off across iterations. This adaptive mechanism can empirically mitigate mismatches between the surrogate objective and the underlying data distribution, and can further tailor the optimization criterion to downstream performance measures.
} 
Furthermore, even when the original objective is an accurate representation of the task, the trainable nature of this methodology enables the design of {\em alternative objectives} that optimize for downstream performance measures. 
For example, it was shown in \cite{khobahi2021lord} that when optimization-based equalization is followed by symbol detection in quantized \ac{mimo} systems, learning the objective parameters allows the unfolded optimizer to adapt its criterion such that its outputs are most beneficial for the subsequent detection stage.

Beyond coping with surrogate objectives and latency constraints, this added abstractness also enables a form of {\em robust optimization via robust training}. 
Since the learned objective parameters are tuned from data, one can apply established robust training techniques directly to the unfolded optimizer, without needing to analytically derive new robust formulations. 
This property allows unfolded optimizers to handle not only noisy or uncertain inputs but also deliberately crafted adversarial perturbations~\cite{sofer2025unveiling}. 

While the gains discussed thus far relate primarily to conventional iterative optimization, learning objective parameters also brings forth notable advantages compared to black-box \acp{dnn}. 
\textcolor{NewColor}{Learning objective parameters preserves the algorithmic structure of the unfolded architecture, allowing one to explicitly identify the objective and solver hyperparameters associated with each iteration. This brings forth some level of interpretability (\ref{itm:interpretability}) which is structural here rather than global. In particular, when objectives vary across layers without explicit constraints tying them to a single underlying problem, the resulting network executes a sequence of optimization-inspired transformations. Moreover}, the existence of an exact parameter configuration corresponding to the original non-learned optimizer provides a principled and effective {\em initialization scheme}. 
Initializing the unfolded model with the parameters of the original solver ensures that training begins from a well-understood, interpretable starting point and often leads to faster convergence and improved generalization. 

\textcolor{NewColor}{Finally, recent works have shown that in some cases, the learning objective parameters  can give rise to architectures resembling modern transformer networks. For instance, \cite{yu2023white} derives self-attention-like operations by unrolling an optimization problem associated with sparse rate reduction, where the parameters of the underlying representation model are learned. Similarly, \cite{do2024interpretable} obtains lightweight transformer architectures by unrolling a graph-based signal interpolation problem, in which the graph structure (i.e., edge weights) is learned from data. These examples highlight how learning the parameters of the optimization objective can induce rich, interpretable neural modules.}

\subsection{Learning Correction Term}
\label{ssec:Unfolding_corr}

While classical iterative methods often come with convergence guarantees and an interpretable update map, their practical performance may be hindered by slow convergence when only a limited number of iterations can be executed. 
As the number of layers $K$ is typically small in the unfolding setting, this limitation becomes critical: the truncated classical iteration may remain far from the fixed point of interest. 
To address this issue, an alternative deep unfolding paradigm augments the classical update with a learned correction term that accelerates progress toward a desirable solution while preserving the structural properties of the original method. 
Rather than replacing the update rule, the correction nudges the classical iteration toward trajectories that achieve high performance within a few layers, balancing interpretability with data-driven adaptability.

\paragraph*{Formulation}
Let $\Label^{(k)}$ denote the collection of algorithmic variables at layer $k$, and let $\mathcal{T}$ denote the classical update operator. 
Here, $\mathcal{T}$ can represent the generic iterative updates in~\eqref{eqn:FullGD} and~\eqref{eqn:ProxGD}, or more problem-specific iterations as those in~\eqref{eqn:PGAsteps}.
Thus, we denote the $k$th iteration of the classical optimizer as $\mathcal{T}(\Label^{(k-1)})$.
In the discussed unfolding paradigm, we augment this classical update with a learned correction term that (potentially) depends on both the problem instance $\Input$ and the current iterate
\begin{equation}
    \Label^{(k)} = \mathcal{T}(\Label^{(k-1)}) + \Delta(\Label^{(k-1)}, \Input; \myVec{\theta}^{(k)}),
    \label{eq:correction}
\end{equation}
with $\myVec{\theta}^{(k)}$ denoting the trainable parameters of the correction term $\Delta(\cdot)$ in the $k$-th layer.

A few observations are in order. First, unlike the direct parameterization strategy employed in Section~\ref{ssec:Unfold_hyp}, we do not learn the values of the correction terms themselves, but rather the parameters of a neural module $\Delta(\cdot)$ that maps the problem instance and the current iterate into a suitable correction. Directly learning a distinct correction term for each layer would ignore how the update should adapt to the current location in the optimization landscape and would result in a static, state-agnostic perturbation. In contrast, parameterizing $\Delta(\cdot)$ as a function of both $\Label^{(k-1)}$ and $\Input$ enables corrections that are dynamically tailored to the evolving algorithmic state.

Second, although in~\eqref{eq:correction} we write the trainable parameters as $\myVec{\theta}^{(k)}$ to emphasize the possibility of layer-specific parameterization, these parameters need not be layer-dependent. One may instead use a shared parameter set $\myVec{\theta}$ across all layers, whereby the resulting correction still varies with $k$ due to the dependence of $\Delta(\cdot)$ on the evolving iterate $\Label^{(k-1)}$. This parameter-sharing scheme often improves gradient flow during training and enables the unfolded architecture to be evaluated with a different number of layers at inference time, without retraining the model that generates the correction terms.

The form of $\Delta$ may vary significantly across applications, including additive residual mappings, multiplicative reweighting, or corrections applied only to specific subsets of variables. Importantly, the aim is not to replace $\mathcal{T}$ with a learned update rule, but to bias its trajectory using data while retaining the interpretability and asymptotic properties of the original iteration.
A representative example of this methodology is the unfolded WMMSE (UWMMSE) algorithm~\cite{chowdhury2021unfolding, chowdhury2023deep}, which is described in the box on Page~\pageref{Box:LearnCorrExm}.

\paragraph*{Gains}
Learning a correction term offers a principled way to enhance classical iterative methods while preserving their algorithmic structure. By refining (rather than replacing) the update $\mathcal{T}$, this approach maintains a strong inductive bias toward the original optimization task, preserves interpretability of intermediate computations, and enables performance improvements even when the classical iteration is truncated to a few layers. At the same time, the learned correction introduces robustness to modeling inaccuracies and surrogate formulations, as it can compensate for mismatches between the mathematical objective and the true task of interest.

From a learning perspective, training a neural module to generate corrections often results in greater stability and data efficiency than directly learning algorithmic variables, since updates are applied to meaningful iterates rather than produced from scratch. Moreover, the correction network can be lightweight and its parameters may be shared across layers, enabling deployment with a different number of iterations at inference time without retraining. While substantial changes in the objective parameters $\ObjParam$ may require retraining, this methodology remains adaptable across problem instances and retains computational efficiency due to the reliance on a small number of structured forward passes.

\begin{tcolorbox}[float*=t,
    width=\boxWidth,
    toprule = 0mm,
    bottomrule = 0mm,
    leftrule = 0mm,
    rightrule = 0mm,
    arc = 0mm,
    colframe = myblue,
    colback = mypurple,
    fonttitle = \sffamily\bfseries\large,
    title = Unfolded WMMSE with Learned Correction Term]
\label{Box:LearnCorrExm}

We illustrate the learning of a correction term through the unfolded weighted minimum mean-square error (UWMMSE) architecture for power allocation in interference networks. For a given channel realization $\Input = \myMat{H}$ with $M$ transmitter–receiver pairs, the classical WMMSE algorithm iteratively updates $\Label^{(k)} \triangleq \big( \mathbf{u}^{(k)}, \mathbf{w}^{(k)}, \mathbf{v}^{(k)} \big)$, with $\mathbf{u}^{(k)}, \mathbf{w}^{(k)}, \mathbf{v}^{(k)} \in \mathbb{R}^M$, according to closed-form expressions. 
Focusing on a generic user $i$, the classical weight update takes the scalar form
\[
    w^{(k)}_{i,\mathrm{class}}
    = \frac{1}{1 - u^{(k)}_i h_{ii} v^{(k-1)}_i},
\]
where $h_{ii}$ is the direct channel gain for user $i$.

In UWMMSE, this scalar update is refined by an \emph{affine correction} that is learned from data. Specifically, for each user $i$ and layer $k$,
\begin{equation}
    w^{(k)}_i = a^{(k)}_i\, w^{(k)}_{i,\mathrm{class}} + b^{(k)}_i,
    \label{eq:uwmmse_affine}
\end{equation}
so that the classical update is first computed and then scaled and shifted by learned coefficients. Collecting these coefficients across all users yields vectors $\mathbf{a}^{(k)} = \big[a^{(k)}_1,\ldots,a^{(k)}_M\big]^\top$, $\mathbf{b}^{(k)} = \big[b^{(k)}_1,\ldots,b^{(k)}_M\big]^\top$,
which are produced by two permutation-equivariant graph neural networks
\[
    \mathbf{a}^{(k)} = \Psi_a(\Label^{(k-1)}, \myMat{H}; \myVec{\theta}_a),
    \qquad
    \mathbf{b}^{(k)} = \Psi_b(\Label^{(k-1)}, \myMat{H}; \myVec{\theta}_b).
\]
The GNN parameters $\myVec{\theta}_a$ and $\myVec{\theta}_b$ are shared across users, so the number of trainable parameters does not grow with $M$. At the same time, the output vectors $\mathbf{a}^{(k)}$ and $\mathbf{b}^{(k)}$ adapt in size to the number of transceivers, allowing the same unfolded architecture to operate on networks of different dimensions without architectural changes.
This construction is a concrete instance of the learned correction term in \eqref{eq:correction}: the operator $\mathcal{T}$ implements the classical WMMSE update, while the correction acts only on the $\mathbf{w}$-block via the affine refinement \eqref{eq:uwmmse_affine}. 
\end{tcolorbox}

\subsection{\ac{dnn} Inductive Bias}
\label{ssec:Unfolding_Ind}
The final design methodology within the deep unfolding framework is the one that most closely resembles black-box \acp{dnn}. 
Rather than preserving the explicit step-by-step operation of an iterative solver, this approach constructs a deep neural architecture whose inductive bias stems from the information flow and structure of iterative optimization algorithms that are well-suited to the considered task. 
Consequently, the model is not based on an off-the-shelf network architecture from computer vision or natural language processing, but rather on one that is {\em inspired by}, rather than directly derived from, a classical optimization process. 
This design paradigm offers the most abstract and flexible form of deep unfolding, typically resulting in highly parameterized architectures that provide strong representational power at the cost of reduced interpretability. 

\paragraph*{Formulation}
As in the previous methodologies, the starting point here is an iterative optimizer operating for a fixed number of iterations~$K$. 
However, in this approach, the conventional computational steps of the iterative solver are not preserved explicitly. 
Instead, the mapping carried out by each iteration~$k$, which traditionally updates $\Label^{(k)}$ based on the previous estimate~$\Label^{(k-1)}$ and the input~$\Input$, is implemented through a dedicated \ac{dnn} parameterized by~$\dnnParam_k$. 
Each such neural module may either augment certain components of the original iteration, or substitute the entire iteration altogether. 
The resulting model thus forms a modular deep architecture in which the number of \ac{dnn} modules corresponds to the number of unfolded iterations, while the operations within each layer are learned rather than analytically defined. 
A representative example of this design methodology is the {\em DeepSIC} architecture for multiuser \ac{mimo} symbol detection of~\cite{shlezinger2020deepsic}, described in the box entitled {\em DeepSIC} on Page~\pageref{Box:LearnIndExm}.

\begin{tcolorbox}[float*=t,
    width=\boxWidth,
	toprule = 0mm,
	bottomrule = 0mm,
	leftrule = 0mm,
	rightrule = 0mm,
	arc = 0mm,
	colframe = myblue,
	colback = mypurple,
	fonttitle = \sffamily\bfseries\large,
	title = DeepSIC]	
	\label{Box:LearnIndExm} 
 Multiuser \ac{mimo} symbol detection considers the recovery of $U$ constellation symbols $\Label_1,\ldots,\Label_U$ from an observed vector $\Input$. The iterative soft interference cancellation algorithm of \cite{choi2000} has each iteration $k$  gradually refines the probability mass function estimate $\myVec{p}_u^{(k)}$ for each $u \in \{1,\ldots, U\}$. This is achieved by assuming a linear Gaussian relationship where $\Input = \sum_{u=1}^U \myVec{h}_u \Label_u + \myVec{w}$ with known channels $\{\myVec{h}_u\}$, where $\myVec{w}$ is Gaussian noise, based on which the previous soft estimates of the interfering symbols $\{\myVec{p}_{j}^{(k-1)}\}_{j\neq u}$ are employed to subtract the interference contribution from $\Input$ and estimate the probability mass function $\myVec{p}_u^{(k)}$. Once convergence is achieved, each symbol is obtained by taking the argmax from its corresponding soft estimate. 

\begin{wrapfigure}{r}{0.55\linewidth} 
    \centering
    \vspace{-0.3cm}
    \includegraphics[width=\linewidth]{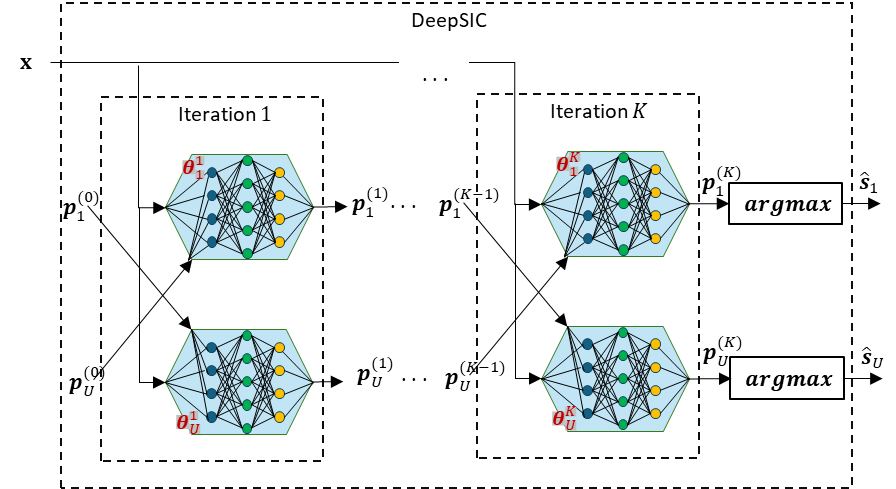} 
    \caption{DeepSIC illustration, with $\dnnParam$ marked in \textcolor{red}{red} } 
    \label{fig:DeepSIC}
\end{wrapfigure}
 The DeepSIC algorithm of \cite{shlezinger2020deepsic}, illustrated in Fig.~\ref{fig:DeepSIC}, unfolds iterative soft interference cancellation into $K$ iterations, and replaces each mapping from $\{\myVec{p}_{j}^{(k-1)}\}_{j\neq u}$ and $\Input$ into $\myVec{p}_u^{(k)}$ with a compact \ac{dnn} parameterized by $\dnnParam_u^k$. The \ac{dnn} has  a softmax output layer to output a probability mass function, while learning to cancel interference from data without any explicit statistical modeling. 
\end{tcolorbox}

\paragraph*{Gains}
This design methodology yields \acp{dnn} that substantially depart from their original optimization counterparts, inheriting the expressive power and abstractness of deep learning while maintaining a structural link to principled optimization. 
Its immediate gain lies in its ability to learn inference mechanisms that no longer depend on simplified (\ref{itm:Relaxation}) or approximate (\ref{itm:Objective}) mathematical objectives. 
For instance, while the soft interference cancellation algorithm that motivates DeepSIC is fundamentally derived under linear Gaussian \ac{mimo} channel assumptions, its deep unfolded version can learn to operate effectively in nonlinear, correlated, or otherwise mismatched channel conditions without the need for explicit analytical reformulation.

Another notable benefit, albeit less direct, is the potential reduction in latency compared to conventional iterative solvers (\ref{itm:Latency}). 
Although the number of iterations~$K$ remains small and fixed, each iteration now involves evaluating a compact \ac{dnn}, which, once trained, can be executed rapidly and is highly amenable to parallelization. 
In many cases, this leads to significantly faster inference than classical optimization iterations that require slow and intensive operations such as matrix inversions or factorization steps~\cite{revach2023rtsnet}.

While this methodology does not preserve the same level of adaptability (\ref{itm:Adaptability}) and scalability (\ref{itm:Scalability}) as classical iterative optimizers, it remains more flexible and structured than conventional black-box \acp{dnn}. 
For example, although the DeepSIC model requires retraining or adaptation when channel statistics vary, its modular structure, which comprises multiple compact \acp{dnn} that can be individually evaluated, facilitates efficient continual learning and online adaptation~\cite{raviv2023online}. 
Specifically, it allows for rapid single-step continual updates~\cite{gusakov2025rapid}, selective retraining of individual modules~\cite{uzlaner2025async}, or integration with hypernetworks that dynamically generate the parameters of each iteration module in real time~\cite{raviv2025modular}. 
Such hybrid designs enable the architecture to elastically adapt to changing environments, combining the modularity and interpretability of iterative optimizers with the flexibility and learning capacity of \acp{dnn}.

\subsection{Training Unfolded Architectures}
\label{ssec:Unfolding_Train} 

\paragraph*{Differentiability} 
The casting of iterative optimizers as \ac{ml} architectures enables the tuning of their parameters $\dnnParam$ to best match the available data $\mySet{D}$ using gradient-based deep learning techniques. Such approaches require the resulting \ac{ml} model to be end-to-end differentiable. While \acp{dnn} are end-to-end differentiable by design, unfolded architectures arise from iterative optimizers that were not specifically crafted to be differentiable. Fortunately, end-to-end differentiability  typically holds for convex iterative optimizers~\cite{agrawal2021learning}, and can be tackled via dedicated approximations when it does not hold~\cite{craven1986nondifferentiable}. 

\color{NewColor}
\paragraph*{Parameter Tying} 
The different formulations presented in the previous subsections specify how one can convert an iteration of an iterative optimizer into a trainable parameterized \ac{ml} model. In learning these parameters $\dnnParam$, one can choose  whether to tie parameters across iterations or allow iteration-dependent parameterizations. Tying weights preserves a closer correspondence with the original iterative algorithm, reduces the number of trainable parameters, and often enhances stability and generalization.   Alternatively, allowing iteration-specific parameters increases expressivity at the cost of a larger model. Both regimes are fully compatible with the deep unfolding paradigm, and the choice between them reflects a trade-off between model capacity and structural regularization.
\color{black}

\paragraph*{Loss Formulation}
One of the key advantages of deep unfolded \ac{ml} architectures compared to conventional black-box \acp{dnn} is their enhanced training flexibility, which holds regardless of the specific unfolding methodology employed. This flexibility manifests in two main aspects: 
$(i)$ deep unfolded models often support both supervised and unsupervised learning; 
$(ii)$ their structured iterative nature allows the formulation of training losses that account not only for the final output but also for intermediate iteration-wise estimates. These characteristics, formulated next, make deep unfolded architectures highly adaptable to different forms of supervision and enable richer and more stable learning dynamics.

{\bf Supervision Type}: 
A defining feature of deep unfolded optimizers is their origin in iterative optimization solvers. This origin inherently provides a well-defined criterion for assessing the quality of their outputs via the optimization objective $\mySet{L}_{\ObjParam}(\cdot)$. As such, deep unfolded models can be trained in an {\em unsupervised} manner, using the same objective that originally guided the iterative solver as a sample-wise loss function in the empirical risk formulation, i.e., using an empirical risk of the form
\begin{equation}
    \mySet{L}_{\mySet{D}}^{\rm ER}(\dnnParam) = \frac{1}{|\mySet{D}|} \sum_{\Input_t \in \mySet{D}} \mySet{L}_{\ObjParam}\left(\dnnFunc(\Input_t); \Input_t\right). 
    \label{eqn:UnsupUnfold}
\end{equation}
\textcolor{NewColor}{Clearly, unfolded networks can also be trained using alternative \ac{ml} self-supervised criteria, such as data-consistency-based losses or Noise2Noise-type formulations~\cite{janjuvsevic2026self}. The key advantage of deep unfolding is that, in addition to these generic learning strategies, it naturally provides a principled objective function that can serve as a sample-wise training loss.}

When labeled data are available, however, {\em supervised} learning becomes an appealing alternative. In this setting, the loss can be formulated based on some  difference measure $l(\cdot)$ between the model output $\dnnFunc(\Input_t)$ and the target labels $\Label_t$ as in \eqref{eqn:SupervisedLoss}, rather than solely evaluating the optimization objective value. This flexibility enables one to deviate from potential mismatches induced by the surrogate nature of the optimization objective used in model-based inference, while maintaining the interpretability and structured operation of the underlying iterative solver. Naturally, the flexibility in supervision type can also be leveraged via semi-supervised learning, i.e., by combining supervised and unsupervised learning, which is useful when the number of labeled samples is small. 

{\bf Iteration-Wise Loss Formulation}: 
The interpretability of iterative optimizers, and specifically the fact that each iteration yields a progressively refined version of the optimization variable, introduces an additional degree of freedom in training deep unfolded architectures. In conventional black-box \acp{dnn}, a meaningful loss can typically be assigned only to the output of the network, which corresponds to the final decision $\dnnFunc(\Input_t)$. This practice, though natural, can make training challenging: the learning process must jointly optimize a large number of parameters based only on feedback from the output layer, often requiring extensive datasets and suffering from vanishing gradient effects.

The sequential structure of deep unfolded models provides an opportunity to incorporate losses associated with intermediate iterations. To formulate this mathematically, we focus on supervised learning as in \eqref{eqn:SupervisedLoss} with sample-wise loss $l(\cdot)$ and without regularization (i.e., $\lambda=0$). By denoting the trainable parameters of the $k$th iteration as $\dnnParam_k$ \textcolor{NewColor}{(focusing on the generic case where the parameters can change between iteration)}, i.e., $\dnnParam=\{\dnnParam_k\}$, and writing ${\Label}^{(k)}(\Input;\dnnParam_1,\ldots, \dnnParam_k)$ as the output of the $k$th iteration of an unfolded architecture with trainable parameters $\{\dnnParam_k\}$, we can identify three different training loss options:
\begin{subequations}
    \label{eqn:LossTypes}
\begin{enumerate}[label={L\arabic*}]
    \item \label{itm:e2e} {\bf End-to-end loss},  which follows the common deep learning practice and  evaluates the model based on its output, which here is $\dnnFunc(\Input) = {\Label}^{(K)}(\Input;\dnnParam_1,\ldots, \dnnParam_k)$, and thus 
    \begin{equation}
        \mySet{L}_{\mySet{D}}^{\rm ER}(\dnnParam) = \frac{1}{|\mySet{D}|} \sum_{(\Input_t,\Label_t) \in \mySet{D}} l\left( {\Label}^{(K)}(\Input_t;\dnnParam_1,\ldots, \dnnParam_K), \Label_t\right).    
    \label{eqn:LossTypesE2E}
    \end{equation}
    \item \label{itm:multiiter} {\bf Multi-iteration loss}, that aggregates the losses from multiple iterations, typically as a weighted combination that balances the importance of early and late stages, i.e., 
    \begin{equation}
        \mySet{L}_{\mySet{D}}^{\rm ER}(\dnnParam) = \frac{1}{|\mySet{D}|} \sum_{(\Input_t,\Label_t) \in \mySet{D}} \sum_{k=1}^K \alpha_k \cdot l\left( {\Label}^{(k)}(\Input_t;\dnnParam_1,\ldots, \dnnParam_k), \Label_t\right).    
    \label{eqn:LossTypesmultiiter}
    \end{equation}
    In \eqref{eqn:LossTypesmultiiter}, $\{\alpha_k\}$ represent the weight assigned to each iteration, with a candidate setting proposed in \cite{samuel2019learning} sets $\alpha_k=\log(1+k)$ to gradually assign more weight to later iterations. 
    \item \label{itm:sequential} {\bf Sequential loss}, that follows a form of progressive learning,  gradually trains the parameters of successive iterations based on the empirical risk
        \begin{equation}
        \mySet{L}_{\mySet{D}}^{\rm ER}(\dnnParam_k) = \frac{1}{|\mySet{D}|} \sum_{(\Input_t,\Label_t) \in \mySet{D}}  l\left( {\Label}^{(k)}(\Input_t;\dnnParam_1,\ldots, \dnnParam_k), \Label_t\right).    
    \label{eqn:LossTypessequential}
    \end{equation}
    Sequential training gradually trains the parameters of each iteration $k=1,2,\ldots, K$, where typically the parameters of iterations $1,\ldots, k-1$ are fixed when training $\dnnParam_k$ as proposed in \cite{shlezinger2020deepsic}. 
\end{enumerate}
\end{subequations}
As opposed to the standard end-to-end approach \ref{itm:e2e}, the iteration-wise formulations \ref{itm:multiiter}-\ref{itm:sequential} both  leverage the interpretable intermediate variables produced by the unfolded structure to improve convergence behavior, enhance gradient flow, and promote consistent refinement across iterations. Among the two, \ref{itm:sequential} is more geared towards preserving the operation of conventional descent-based optimization, where each iteration aims to improve upon its preceding ones in terms of the loss $l(\cdot)$, while \ref{itm:multiiter} allows some flexibility for intermediate iterations to deviate from such operation.


\section{Theoretical Foundations}
\label{sec:Theoretical}
\begingroup

\newcommand{\paren}[1]{ \left( #1 \right) }
\newcommand{\brac}[1]{ \left[ #1 \right] }
\newcommand{\bracset}[1]{ \left\{ #1 \right\} }

\newcommand{\Rcal}{\mySet{R}}
\newcommand{\Lcal}{\mySet{L}}
\newcommand{\Dcal}{\mySet{D}}
\newcommand{\Ecal}{\mySet{E}}
\newcommand{\Rad}{\mathfrak{R}}

\newcommand{\xv}{\myVec{x}}
\newcommand{\sv}{\myVec{s}}
\newcommand{\thetav}{\myVec{\theta}}

\newcommand{\norm}[1]{ \left\lVert #1 \right\rVert }
\newcommand{\abs}[1]{ \left\lvert #1 \right\rvert }

\newcommand{\Expt}[2][]{\mathbb{E}_{#1}\brac{ #2 }}
\newcommand{\Prob}[1]{\mathbb{P}\paren{ #1 }}

So far, we have introduced different forms of converting iterative optimizers used for inference into trainable \ac{ml} architectures via deep unfolding. One of the core gains of designing \ac{ml} models based on principled and mathematically grounded optimization solvers is the ability to preserve some of the theoretical soundness and performance guarantees of such classical algorithms. In this
 section we present recent theoretical advances that provide rigorous analysis and performance guarantees for deep unfolded optimizers. While the theoretical backing of deep unfolded optimizers is still an area of ongoing research, our aim is to provide both practical understanding and theoretical justification for the design and application of deep unfolding architectures.

 Deep unfolded models can be viewed from two complementary angles: either as a parameterized optimization algorithm that learns its parameters from data; and as a highly structured neural network with its architecture inspired from an iterative algorithm. Motivated by these viewpoints, existing theoretical characterization of unfolded optimizers include: $(i)$ convergence and rates of unfolded optimizers toward solutions of the target inference/optimization problems; $(ii)$ generalization guarantees and error bounds of deep unfolded networks; and $(iii)$ training-time optimization guarantees, which analyze the dynamics of learning the parameters and identify conditions under which training is stable and convergent. We organize our review based on this division and highlight representative results under each lens.
 For broader perspectives that complement (but are not central to) our focus, readers may consult several surveys \cite{monga2021algorithm,scarlett2022theoretical,chen2022learning}, which cover topics such as conceptual links to classical algorithms \cite{monga2021algorithm}, theoretical results on general deep-learning-based algorithms for inverse problems \cite{scarlett2022theoretical}, as well as relation to learning-to-optimize approaches \cite{chen2022learning}.
 
 \subsection{Convergence Guarantees of Deep Unfolded Optimizers}
 
  \begin{tcolorbox}[float*=t,
 	width=\boxWidth,
 	toprule = 0mm,
 	bottomrule = 0mm,
 	leftrule = 0mm,
 	rightrule = 0mm,
 	arc = 0mm,
 	colframe = myblue,
 	colback = mypurple,
 	fonttitle = \sffamily\bfseries\large,
 	title = Parameter Coupling and Convergence Rates in LISTA and Its Variants]	
 	\label{Box:theory:convergence_rate} 
 	Consider the sparse signal recovery problem arising from \eqref{eqn:InvesObj1} and assume $\myMat{\Psi}=\myMat{I}$ for simplicity. Let the observation be
 	$\myVec{x}=\myMat{H}\myVec{s}_\star+\myVec{e}$, where $\myVec{s}_\star$ is the ground-truth sparse signal and $\myVec{e}$ is observation noise. We estimate $\myVec{s}_\star$ using the $k$-th layer output $\myVec{s}^{(k)}$ of the LISTA network \eqref{eqn:LISTA}. 
 	
 	Assuming that $\lVert \myVec{s}_\star \rVert_{\infty}\leq B$, $\lVert{\myVec{s}_\star}\rVert_0 \leq \eta$, and $\lVert{\myVec{e}}\rVert_1\leq \sigma$, Chen et al. \cite{chen2018theoretical} established the following: 	
 	\begin{enumerate}
 		\item If $\sigma=0$ and $\myVec{s}^{(k)}\to\myVec{s}_\star$ as $k\to \infty$, then under mild conditions the parameters must satisfy
 		\begin{equation}\label{eq:CP_limit}
 			\myMat{W}_k^2-(\myMat{I}-\myMat{W}_k^1 \myMat{H})\to 0,\qquad \beta_k\to 0 .
 		\end{equation}
 		
 		\item Building on \eqref{eq:CP_limit}, the authors proposed a coupled-parameter (CP) technique to reduce the number of learnable parameters by enforcing $\myMat{W}_k^2 \triangleq \myMat{I}-\myMat{W}_k^1 \myMat{H}$. They proved that, when $\eta$ is sufficiently small, there exists $\left\{ \myMat{W}_k^1, \beta_k \right\}_{k=1}^\infty$ and constants $c,C>0$ that guarantee for all possible $(\myVec{s}_\star,\myVec{e})$:
 		\[\left\lVert \myVec{s}^{(k)} - \myVec{s}_\star \right\lVert_2 \leq \eta B\exp(-ck)+C\sigma,\quad \forall k=1,2,\cdots.\]
 	\end{enumerate}
 	
 	Furthermore, it was shown in \cite{liu2019alista}  that for  LISTA-CP, the weight matrices $\myMat{W}_k^1$ need not be learned from data; it suffices to take them proportional to a matrix that minimizes the mutual coherence of $\myVec{H}$.\footnote{The mutual coherence optimization problem is nonconvex, thus it is difficult to compute global minimizers in general.} 
 	Under reasonable conditions, it is shown that this analytical choice (the resultant design is termed ALISTA) yields linear convergence, while the convergence rate of LISTA cannot be faster than linear, thus leading to a near-optimality architecture with better parameter efficiency. 
 	Moreover, by introducing a momentum term to ALISTA and instance-adaptive hyperparameter scheduling,  the resulting unrolled optimizer can be guaranteed to achieve superlinear convergence under appropriate assumptions\cite{chen2021hyperparameter}.
 	
 	More recently, several studies have extended these ideas to broader sparse-recovery settings. For example, Yang et al. \cite{yang2025deep} analyze an unfolded architecture for multiple measurement vectors sparse recovery and derive CP-type relations and linear convergence guarantees akin to \cite{chen2018theoretical}, together with analytic expressions for certain weights analogous to \cite{liu2019alista}. Yang et al. \cite{yang2025improving} unfold a proximal-gradient method for the Elastic Net, and establish CP-type relations and linear convergence.
 \end{tcolorbox}

By viewing deep unfolded models as learned optimization algorithms, one can analyze its convergence to the solution of the target inference/optimization problem and the attainable convergence rates from an optimization-theoretic standpoint.
Early studies \cite{chen2018theoretical,liu2019alista,chen2021hyperparameter} mostly focused on concrete inference/optimization tasks (e.g., sparse recovery) and treated particular unfolded architectures (most notably LISTA and its variants) as optimization algorithms whose depth plays the role of iteration count. Using this framework, the following types of results are typically revealed:
\begin{enumerate}
	\item {\bf Parameters coupling}. Leveraging available model knowledge, one can often show that convergence of the unfolded optimizer implies certain dependency among the learnable parameters, which yields unfolding architectures with fewer trainable parameters without degrading the convergence rate.
	
	\item {\bf Existence of parameters that ensure fast (linear or even superlinear) convergence}. Such results justify the unfolding architecture by showing that it can represent\footnote{Importantly, these results do not guarantee that standard training procedures will actually discover those optimal parameter values.} the desired update rule achieving much faster convergence than hand-crafted optimizers with far fewer parameters than generic \acp{dnn}.
\end{enumerate}
We highlight representative results along these two themes for LISTA-type unfolded optimizers in the boxed entitled ``Parameter Coupling and Convergence Rates in LISTA and Its Variants" on Page~\pageref{Box:theory:convergence_rate}.

Beyond problem-specific analyses, there is a complementary line of theory that targets more general unfolded architectures. For example, Heaton et al. \cite{heaton2023safeguarded} propose architecture-agnostic safeguarding mechanisms to ensure convergence of unfolded fixed point algorithms. Hadou et al. \cite{hadou2024robust} propose imposing descending constraints in training general unfolded optimizers, and establish high-probability convergence guarantees and exponential decrease of the objective gradient norm under certain assumptions. The strength of these approaches lies in their generality, as they are not tied to a particular optimization problem or unfolding architecture; the trade-off is that, by abstracting away architectural details, the conclusions are typically either asymptotic (e.g., establishing only that the unfolded optimizer converges as depth increases without quantifying the explicit dependence on the number of layers), or require stronger assumptions that are difficult to verify in practice.
 
 \subsection{Generalization Error Bounds of Unfolded Networks}

As a special class of neural architectures, unfolded networks can be analyzed through the lens of statistical learning theory. A central question is whether a network that achieves very small training error will also incur small loss on unseen data. Tackling this question involves characterizing the generalization error, i.e., the gap between the expected loss and the empirical loss, and how this gap depends on the number of parameters, the network depth (i.e., the number of unrolled iterations), and the number of training samples.

 \begin{tcolorbox}[float*=t,
 	width=\boxWidth,
 	toprule = 0mm,
 	bottomrule = 0mm,
 	leftrule = 0mm,
 	rightrule = 0mm,
 	arc = 0mm,
 	colframe = myblue,
 	colback = mypurple,
 	fonttitle = \sffamily\bfseries\large,
 	title = Generalization Error Bounds of Unfolded Networks for Sparse Recovery]	
 	\label{Box:theory:generalization_error}
 	Consider the sparse recovery problem (\ref{eqn:InvesObj1}) and the unfolded network described in \eqref{eq:generalization_setup}. Behboodi et al. \cite{behboodi2022compressive} assumed that the dictionary $\myMat{\Psi}\in\mathbb{R}^{N\times N}$ in \eqref{eqn:InvesObj1} is an unknown orthogonal matrix, and deploy an ISTA-like unfolded network to jointly learn the dictionary and recover the signal. The learnable dictionary matrix $\myMat{\Phi}\in\mySet{R}^{N\times N}$, assumed orthogonal, is shared across all layers. Using Dudley's inequality, the authors transform the problem of upper-bounding $\Rad_{|\Dcal|}$ in \eqref{eq:Rademacher_complexity} into upper-bounding covering numbers of sample-output matrix class 
 	\begin{equation*}
 		\mathcal{M}\triangleq\bracset{ \begin{bmatrix}
 				\sv^{(K)}(\xv_1;\myMat{\Phi}) & \cdots & \sv^{(K)}(\xv_{\abs{\Dcal}};\myMat{\Phi})
 		\end{bmatrix}  \;\middle\vert\; \myMat{\Phi}\in \mySet{R}^{N\times N} ~\text{is orthogonal} },
 	\end{equation*}
 	thereby deriving a generalization bound. Schnoor et al. \cite{schnoor2023generalization} extended this framework to allow a more general architecture with arbitrary parameter sharing across per-layer weight matrices. With fixed threshold parameters, their high-probability generalization bound can be simplified to
 	\begin{equation*}
 		\Lcal(\thetav) \leq \Lcal^{\rm ER}_{\Dcal}(\thetav) + \mathcal{O} \paren{ \sqrt{ \frac{P\log(K)}{\abs{\Dcal}} } },
 	\end{equation*} 	
 	where $P$ is the dimension of learnable parameters. Notably, the dependence of this generalization bound on depth $K$ is logarithmic, rather than exponential as in some earlier bounds for generic neural networks (derived in classification problems, however).
 	
 	Shultzman et al. \cite{shultzman2023generalization} considered the case $\myMat{\Psi}=\myMat{I}$ and studied the expected supremum of generalization gap of two unfolded networks derived from ISTA and ADMM. A key observation of \cite{shultzman2023generalization} is that the soft-thresholding nonlinearity in unfolded networks can reduce the Rademacher complexity, which constitutes the crucial tool for bounding the generalization error. It is shown that by controlling the norms of the weight matrices, a nonincreasing generalization error as a function of the network's depth is achievable, yielding smaller bounds than those for ReLU networks. The authors also study the estimation error $\Lcal(\theta_{\Dcal}) - \inf_{\thetav} \Lcal(\thetav)$, where $\thetav_{\Dcal}\triangleq \arg \min_{\thetav}\Lcal^{\rm ER}_{\Dcal}(\thetav)$, via local Rademacher complexity (defined by replacing the constraint $\thetav\in\Theta$ in \eqref{eq:Rademacher_complexity} by a neighborhood of $\thetav_{\Dcal}$). The authors prove an $\mathcal{O}(1/|\Dcal|)$ decay for the estimation error (instead of the $\mathcal{O}(1/\sqrt{|\Dcal|})$ scaling typical of uniform generalization bounds) and again show that the estimation error bounds for unfolded networks are smaller than those for ReLU networks (though the gap shrinks as $|\Dcal|$ grows).
 \end{tcolorbox}

One line of work \cite{behboodi2022compressive, schnoor2023generalization, shultzman2023generalization} targets specific unfolded architectures and compares their generalization bounds with those of more generic networks (e.g., ReLU networks), thereby offering an explanation for the empirical advantages of unfolding architectures observed in practice. Consider, for example, the sparse signal recovery problem that arises from \eqref{eqn:InvesObj1}. We estimate the ground-truth signal $\sv$ using the $K$th-layer output $\sv^{(K)}(\xv;\thetav)$ of an unfolded network, where the network is trained on a dataset $\Dcal\triangleq\left\{ (\myVec{x}_t, \myVec{s}_t) \right\}_{t=1}^{|\mySet{D}|}$ i.i.d. drawn from a data-generating distribution $p_{\xv,\sv}$. Define the expected and empirical risks as
\begin{equation}\label{eq:generalization_setup}
	\Lcal(\thetav)\triangleq \Expt[(\xv,\sv)\sim p_{\xv,\sv}]{l\paren{ \sv^{(K)}(\xv;\thetav), \sv } } ,	\quad \Lcal_{\Dcal}^{\rm ER}(\thetav) \triangleq \frac{1}{|\Dcal|} \sum_{(\xv_t,\sv_t) \in \Dcal} l\paren{ \sv^{(K)}(\xv_t;\thetav), \sv_t} .
\end{equation}
To assess performance on unseen data, we study the generalization gap $\Lcal(\thetav)-\Lcal^{\rm ER}_{\Dcal}(\thetav)$. Since $\Lcal^{\rm ER}_{\Dcal}(\thetav)$ is random, we typically consider high-probability bounds of the generalization gap or bounds on the expected supremum $\Expt[\Dcal]{\sup_{\thetav} \paren{\Lcal(\thetav) - \Lcal^{\rm ER}_{\Dcal}(\thetav)}}$. Both types of bounds can be controlled by the Rademacher complexity defined as~\cite[Ch. 26]{shalev2014understanding}
\begin{equation}\label{eq:Rademacher_complexity}
	\Rad_{\abs{\Dcal}}\triangleq \Expt[\Dcal,\bracset{\epsilon_t}_{t=1}^{\abs{\Dcal}}]{ \sup_{\thetav\in{\Theta}} \frac{1}{\abs{\Dcal}} \sum_{t=1}^{\abs{\Dcal}} \epsilon_t \cdot  l\paren{\sv^{(K)}(\xv_t;\thetav), \sv_t  } },
\end{equation}
where $\Theta$ is the set of possible parameter values, $\bracset{\epsilon_t}_{t=1}^{\abs{\Dcal}}$ are i.i.d. Rademacher variables with $\Prob{\epsilon_t=1}=\Prob{\epsilon_t=-1}=\frac{1}{2}$. Hence, bounding the generalization gap reduces to bounding $\Rad_{\abs{\Dcal}}$. We collect several representative results for sparse-recovery–oriented unfolded networks, together with key proof ideas, in the box titled ``Generalization Error Bounds of Unfolded Networks for Sparse Recovery" on Page~\pageref{Box:theory:generalization_error}.  While these are stated for particular architectures, the analysis techniques often extend to many other unfolded designs. 

A few caveats are in order. First, generalization bounds are commonly derived from the entire hypothesis function class; the network obtained after training is just one element of that class and may in practice enjoy a smaller generalization gap than the high-probability upper bound suggests\footnote{Conversely, estimation-error bounds studied in, e.g., \cite{shultzman2023generalization}, that focus on the empirical risk minimizer presume exact minimization of the training loss, which may not be achieved by a practical optimizer.}. Second, present-day Rademacher complexity analyses are largely distribution-agnostic: they do not exploit the detailed structure of the data-generating process. In signal processing and inverse problems, however, the observation model often induces simpler joint distributions and additional dependencies that could—at least in principle—be leveraged to obtain tighter, problem-dependent bounds; current tools only partially capture this structure.

A complementary thread does not assume any specific architectural details, and instead investigates generalization (or overall learning performance) under broad unfolding frameworks. For example, Chen et al. \cite{chen2020understanding} study an objective-parameter–learning–type unfolded optimizer for quadratic optimization and develop relationships connecting convergence rate, stability, and generalization error of the unfolded optimizer. Sucker and Ochs \cite{sucker2023pac-bayesian} employ PAC-Bayes theory to derive a PAC-style generalization bound for a general class of learned optimizers. Hadou et al. \cite{hadou2024robust} analyze unfolded networks trained under descending constraints and prove robustness to distribution shifts. These results are appealingly general, yet they typically do not furnish explicit finite-sample scalings that make the dependence on sample size, depth, or other architectural parameters fully transparent.
 
\subsection{Optimization Guarantees for Unfolded Network Training}

In practice, the performance of an unfolded network is ultimately determined by its training procedure. Yet, training a \ac{dnn} (e.g., based on \eqref{eqn:SupervisedLoss}), is a nonconvex optimization problem: how can we ensure that the learner attains sufficiently small training error rather than getting trapped in a high-error local minimum? Recent advances in the dynamics of neural network training have been leveraged to obtain optimization guarantees for specific unrolled architectures. We highlight two representative lines of work.

Shah et al. \cite{shah2024optimization} studied optimization guarantees of LISTA and an unfolding of ADMM with learned objective parameters for sparse recovery. Using a neural tangent kernel analysis in the overparameterized regime (i.e., the number of trainable parameters exceeds the number of samples), \cite{shah2024optimization} derived the following:
\begin{enumerate}
	\item With i.i.d. Gaussian initialization of the weight matrices and sufficient network width (tied to the ambient signal dimension), if the tangent kernel matrix of the unfolded network is well-conditioned (i.e., its minimum eigenvalue is positive), then with high probability the training loss $\Lcal^{\rm ER}_{\Dcal}(\thetav)$ satisfies a Polyak–Łojasiewicz (PL) condition around the initial parameter $\thetav_0$, i.e., for certain $\mu>0$ and $R>0$,
	\begin{equation}
		\norm{\nabla_{\thetav} \Lcal^{\rm ER}_{\Dcal}(\thetav) }^2_2 \geq \mu \paren{ \Lcal^{\rm ER}_{\Dcal}(\thetav) - \inf \Lcal^{\rm ER}_{\Dcal} }, \qquad \forall \thetav \in B(\thetav_0, R).
	\end{equation}
	
	\item By properties of the PL condition together with the local linearization induced by a well-conditioned tangent kernel matrix, it is further shown that if the number of training samples does not exceed an architecture-dependent threshold, then there exists an interpolating minimizer $\thetav_\star\in B(\thetav_0,R)$ with zero training loss, and gradient descent converges to $\thetav_\star$ at a linear rate when initialized at $\thetav_0$.
\end{enumerate}
Similar results hold for standard feedforward \acp{dnn}; moreover, the authors demonstrate by theory and experiments that, under equal parameter budgets, unfolded architectures can interpolate (i.e., ``memorize") more samples than feedforward \acp{dnn} in the same setting. This suggests that unfolded structures are better matched to sparse recovery, rendering them a powerful architectural choice. Notably, the analysis techniques in this work may be applied to broader families of unfolded architectures in the overparameterized regime.

Karan et al. \cite{karan2024unrolled} consider a specialized unrolled AMP with augmented \acp{dnn} architecture for sparse recovery and prove a stronger generalization-type guarantee. The AMP weight matrices are fixed to those of the classical AMP algorithm, while a two-layer neural network is trained to approximate the scalar denoiser (analogous to the thresholding nonlinearity in ISTA). Assuming that the target signal follows a product (separable) prior across coordinates, it is shown that, under suitable conditions and after sufficiently many steps of online gradient descent, the expected estimation error of the learned unrolled network can be driven within any prescribed tolerance of that achieved by Bayes-AMP, i.e., AMP equipped with the exact posterior-mean (MMSE) denoiser matched to the true prior and noise level. Since Bayes-AMP is known to be optimal among a quite general class of algorithms \cite{montanari2024statistically}, the result establishes a generalization guarantee (not merely a training-error guarantee). However, note that this strong result is enabled by the problem structure: the coordinate-wise product prior and properties of AMP make all the coordinates of the unfolded network output asymptotically behave as if they are i.i.d., which simplifies the analysis.

Overall, optimization guarantees for training unfolded networks remain relatively scarce. Important open directions include guarantees beyond overparameterization and for stochastic-gradient training dynamics. We anticipate that recent progress on training dynamics (e.g., implicit bias phenomena \cite{vardi2023implicit}) and on the convergence of SGD \cite{garrigos2023handbook} will lead to a deeper understanding of when and why unfolded networks can be trained reliably.

\endgroup
\section{Comparative Study}
\label{sec:Comparative}
Section~\ref{sec:Unfolding} presented a set of design approaches and concrete methodologies for converting iterative optimizers into deep unfolded \ac{ml} architectures. \textcolor{NewColor}{Many deep unfolded architectures (particularly those incorporating neural modules within each iteration) can be viewed as end-to-end trained \acp{dnn}~\cite{simon2019rethinking, janjuvsevic2022cdlnet} with a specific, algorithm-inspired architecture. Nonetheless, the deep unfolding paradigm itself is broader, being defined by the systematic transformation of an iterative optimization algorithm into a structured trainable model whose intermediate representations retain an explicit algorithmic interpretation.}

\textcolor{NewColor}{
The reviewed deep unfolding methods all originate from iterative optimization, still,} they notably vary in their strengths, requirements, and implications. To highlight the interplay between the approaches and provide an understanding of how practitioners should prioritize one approach over the others, we next provide a comparative study. We divide this study into two parts: we first present a {\em qualitative comparison}, which pinpoints the conceptual differences between the approaches in light of the identified desired properties   and challenges listed in Table~\ref{tab:myProsConsSummary}. We then detail a {\em quantitative study} for a representative scenario of iterative optimization based on the \ac{rpca} objective to capture the regimes in which each approach is expected to be preferable. 

\subsection{Qualitative Comparison}
\label{ssec:Qualitative} 

The different design methodologies 
reviewed in Section~\ref{sec:Unfolding}, 
provide distinct means of converting iterative optimization solvers into trainable \ac{ml} architectures. 
While they share a common foundation in iterative optimization, their degree of abstraction, interpretability, and learning capacity vary notably. 
To provide a unified perspective, we next conduct a qualitative comparative study highlighting the relative strengths and trade-offs. 
The aspects considered in this comparison are derived from the desired properties and challenges 
discussed in Section~\ref{sec:fund}. 
Specifically, the ability of unfolded optimizers to obtain the minimizer of their corresponding objective (i.e., \ref{itm:Optimality}) is discussed in Section~\ref{sec:Theoretical}, and both challenges \ref{itm:Objective} and \ref{itm:Relaxation}, while originating from different reasons, indicate that the model should be robust to mismatched and approximated objectives. Accordingly, the aspects inspected are
including: 
{\em Adaptability}~(\ref{itm:Adaptability}); 
{\em Interpretability}~(\ref{itm:interpretability}); 
{\em Scalability}~(\ref{itm:Scalability}); 
{\em Abstractness}~(\ref{itm:Objective}–\ref{itm:Relaxation}); 
{\em Latency}~(\ref{itm:Latency}); and 
{\em Complexity}~(\ref{itm:Complexity}). 
The comparison detailed below is summarized in Table~\ref{Tbl:Comparison}.

\begin{table*}
\centering 
\setlength{\tabcolsep}{2pt} 
\renewcommand{\arraystretch}{1.5}
{\scriptsize
\begin{tabular}{|p{1.5cm}|p{2.3cm}|p{2.2cm}|p{2.1cm}|p{2.4cm}|p{2.1cm}|p{2.5cm}|}
 \hline
 {{\bf Method}}  & {\bf Adaptability} \ref{itm:Adaptability} & {\bf Interpretability} \ref{itm:interpretability} & {\bf Scalability} \ref{itm:Scalability} & {\bf Abstractness} \ref{itm:Objective}-\ref{itm:Relaxation} & {\bf Latency} \ref{itm:Latency} & {\bf Complexity} \ref{itm:Complexity} \\ \hline \hline      
  
   Iterative optimizer  & Adaptive to different settings of $\ObjParam$ \cellcolor[HTML]{AAFDB4}         & Fully interpretable\cellcolor[HTML]{AAFDB4}          & Inherently scalable \cellcolor[HTML]{AAFDB4}        &  Relies on accurate task description using a tractable $\mySet{L}_{\ObjParam}$\cellcolor[HTML]{FF9595}     & Requires multiple iterations \cellcolor[HTML]{FF9595}  & Depending on the specific iteration operation \cellcolor[HTML]{FFEAAD}    \\ \hline

   Learning hyperparameters  & Adaptive to different settings of $\ObjParam$ \cellcolor[HTML]{AAFDB4}         & Fully interpretable\cellcolor[HTML]{AAFDB4}          & Inherently scalable \cellcolor[HTML]{AAFDB4}        &  Relies on accurate task description using a tractable $\mySet{L}_{\ObjParam}$, though can learn some limited deviations in training \cellcolor[HTML]{FF9595}     & Supports rapid implementations via few fixed iterations \cellcolor[HTML]{AAFDB4}  & Depending on the specific iteration operation, though complex operation can be relaxed with approximations whose distortion is mitigated via learning~\cite{avrahami2025deep} \cellcolor[HTML]{AAFDB4}    \\ \hline

   Learning objective parameters  & Requires  (possibly simplified) retraining for adaptivity \cellcolor[HTML]{FFEAAD}         & Partially interpretable, \textcolor{NewColor}{as one can identify the objective and solver hyperparameters associated with each iteration}\cellcolor[HTML]{FFEAAD}          & Requires  (possibly simplified) retraining  to handle data with different dimensions \cellcolor[HTML]{FFEAAD}        &  Can support complex mappings that are based on the structure of the objective  $\mySet{L}_{\ObjParam}$ (though not on its parameters $\ObjParam$) \cellcolor[HTML]{AAFDB4}     & Supports rapid implementations via few fixed iterations \cellcolor[HTML]{AAFDB4}  & Depending on the specific iteration operation, though complex operation can be relaxed with approximations whose distortion is mitigated via learning~\cite{avrahami2025deep} \cellcolor[HTML]{AAFDB4}  \\ \hline

   Learning correction term  & Adaptive to different $\ObjParam$ (though substantial variations may require retraining) \cellcolor[HTML]{AAFDB4}         & Preserves fully interpretable computations original solver while including a black-box \ac{dnn} for internal corrections \cellcolor[HTML]{FFEAAD}          & Requires  (possibly simplified) retraining  to handle data with different dimensions \cellcolor[HTML]{FFEAAD}        &  Relies on task description using a tractable $\mySet{L}_{\ObjParam}$, though can learn to correct some level of errors induced by mismatches \cellcolor[HTML]{FFEAAD}     & Few fixed iterations though \ac{dnn} forward pass may induce additional latency \cellcolor[HTML]{FFEAAD} & Depending on the specific iteration operation, combined with few fixed forward passes of compact \acp{dnn} \cellcolor[HTML]{FFEAAD}    \\ \hline

  \ac{dnn} inductive bias  &  Requires  (possibly simplified) retraining for adaptivity \cellcolor[HTML]{FFEAAD}         & Partially interpretable as each iteration is based on \acp{dnn} though features exchanged between iterations represent refined optimization variables\cellcolor[HTML]{FFEAAD}          & Requires    (possibly simplified) retraining  or hypernetworks~\cite{raviv2025modular}  to handle data with different dimensions \cellcolor[HTML]{FFEAAD}        &  Can learn complex mappings and is invariant of  $\mySet{L}_{\ObjParam}$\cellcolor[HTML]{AAFDB4}     & Few fixed iterations though \ac{dnn} forward pass may induce additional latency \cellcolor[HTML]{FFEAAD}  & Few fixed forward passes of compact \acp{dnn} \cellcolor[HTML]{FFEAAD}    \\ \hline

  End-to-end \ac{dnn}   & Not adaptive \cellcolor[HTML]{FF9595}         & Black-box \cellcolor[HTML]{FF9595}          & 
   Some \ac{dnn} architectures may also support variability in some dimensions, though in general, scalability requires  retraining and architecture change  \cellcolor[HTML]{FF9595}        &  Can learn complex mappings and is invariant of  $\mySet{L}_{\ObjParam}$\cellcolor[HTML]{AAFDB4}     &  Depending on number of layers and \ac{dnn} compactness \cellcolor[HTML]{FFEAAD}  & Typically involves complex \acp{dnn} \cellcolor[HTML]{FF9595}    \\ \hline 

\end{tabular}
} 

    \caption{Qualitative comparison between the considered approaches.}
    \label{Tbl:Comparison}
\end{table*}

\paragraph*{Adaptability}
The various 
methodologies differ substantially in their ability to employ the same learned optimizer across multiple settings characterized by different objective parameters~$\ObjParam$. 
Methods based on {\em learning hyperparameters} and {\em learning correction terms} explicitly incorporate 
$\ObjParam$ into their formulation, thereby inheriting the adaptability of classical iterative solvers. 
Once trained, they can typically be applied to new instances of the same optimization problem. 
However, in the case of learned correction terms, the correctional \ac{dnn} components may need fine-tuning when the operating conditions diverge substantially from those observed during training, as their internal representations are implicitly data-dependent. 
Methods that {\em learn objective parameters} or adopt a {\em \ac{dnn} inductive bias} generally require some degree of retraining when the objective or data distribution changes. 
Still, the adaptation process is often more tractable than in black-box \acp{dnn}, since unfolded architectures preserve modularity and interpretability, which allows selective retraining or even targeted adjustment of a subset of parameters~\cite{uzlaner2025async}, and support low-complexity continual learning~\cite{raviv2023online,gusakov2025rapid}. 
Among the two, learning objective parameters tend to require fewer trainable weights, making adaptation simpler and faster. 

\paragraph*{Interpretability}

Among the examined methodologies, those that fully preserve the operation of the original iterative solver (namely, {\em learning hyperparameters}) offer complete interpretability:
Every computation carried out within each iteration corresponds to a known step of the underlying optimization algorithm, and the intermediate features are 
gradual refinements of the estimated optimization variable. 
Moreover, when learning objective parameters, the learned values can be  linked to the objective being optimized at each iteration, providing \textcolor{NewColor}{some level of interpretability as to} how data-driven adaptation affects the mathematical formulation of the problem, \textcolor{NewColor}{though there is not necessarily a formal relationship between the layer-wise objectives and the original formulation.} 
Methods that integrate neural  components introduce some level of black-box behavior. 
When {\em learning correction terms}, interpretability is largely preserved since the architecture still follows the computational flow of the original iterative solver. 
{\em \ac{dnn} inductive bias} 
departs more significantly from the original solver, as each iteration is implemented through a neural module. 
Nonetheless, even in this highly abstract setting, partial interpretability is maintained: the exchanged features between iterations retain their meaning as progressively refined estimates of the decision variable, a property that can be exploited in training and analysis.

\paragraph*{Scalability}
As in \ref{itm:Scalability}, by scalability we refer to the ability of the learned model to generalize across problems of different dimensions, such as varying signal sizes or system configurations. \textcolor{NewColor}{Specific end-to-end \acp{dnn} may also support dimension variability, though this depends on architectural design, and often the architecture depends on the data dimensions.} 
Unfolded architectures that {\em learn hyperparameters} retain the full  structure of the original iterative optimizer and therefore exhibit strong scalability. 
Once trained, they can be  applied to data of different dimensions without any retraining or structural modification, provided the underlying optimization model remains consistent. 
The remaining methodologies are typically more tied to the data dimensions used during training, as these dimensions often determine their parameterization. 
Nevertheless, their modular and interpretable nature allows more efficient adaptation than  black-box \acp{dnn}. 
In particular, dimensional adjustments can often be achieved through limited retraining or the use of {\em modular hypernetworks} that generate dimension-specific parameters for each component as in \cite{raviv2025modular}.

\paragraph*{Abstractness}
Abstractness, in this context, refers to the ability 
to learn effective inference rules even when the original optimization objective does not accurately represent the true underlying task. 
Among the considered methodologies, {\em learning hyperparameters} offers only limited abstractness, as it retains the exact optimization structure. 
While the trainable nature of its hyperparameters introduces some flexibility compared to the conventional solver, its inference behavior remains tightly coupled to the original objective formulation. 
A similar property holds for {\em learning correction terms}, although the inclusion of neural correction modules introduces additional degrees of freedom. 
The remaining approaches provide increased abstractness by allowing the optimizer to deviate more freely from the original mathematical formulation. 
In {\em learning objective parameters}, each iteration can 
effectively redefine what is being optimized at each step to better align with the data-driven task. 
 {\em \ac{dnn} inductive bias}  achieves the highest degree of abstractness: by implementing each iteration as a \ac{dnn}, the learned mappings are no longer constrained by the structure of the original solver, thus allowing the model to discover highly flexible 
 inference rules directly from data.

\paragraph*{Latency}
In the context of iterative optimization, latency corresponds to the number of iterations or computational layers 
executed sequentially. 
Unfolded architectures based on {\em learning hyperparameters} and {\em learning objective parameters} preserve the computational structure of the original solver but operate with a fixed and predefined number of iterations. 
As such, they can 
achieve low latency by setting the number of iterations to a small value while maintaining high performance through  parameter learning. 
Methods that incorporate neural components may introduce additional latency due to the \ac{dnn} forward pass   at each iteration. 
The degree of additional latency depends on the underlying hardware and implementation, as neural computations can often be 
accelerated using dedicated inference engines. 
Despite this, these approaches remain inherently based on iterative solvers with a small and fixed number of iterations, and thus their overall latency typically remains low compared to traditional optimization methods that require numerous iterations to converge. 

\paragraph*{Complexity}
Complexity captures the number of arithmetic operations required by the model during inference. 
Unfolded optimizers that faithfully inherit the operations of the original iterative solver, such as those based on {\em learning hyperparameters} or {\em learning objective parameters}, exhibit low complexity, as they operate with a limited and fixed number of iterations. 
Moreover, their trainable parameters can be optimized to compensate for approximations introduced when simplifying expensive operations, such as matrix inversions, thereby achieving favorable accuracy–complexity trade-offs~\cite{avrahami2025deep}. 
As for  the \ac{dnn}-aided methodologies, i.e., {\em learning correction terms} and {\em \ac{dnn} inductive bias}, their complexity is primarily determined by the size and structure of the neural networks employed within each iteration. 
Still, the deep unfolding framework facilitates using compact, structured modules, whose number of parameters (and thus their computational complexity) remains significantly lower than that of end-to-end black-box \acp{dnn}.

%
%
\subsection{Quantitative Comparison}
\label{ssec:Quantitative}

To complement the theoretical and qualitative insights discussed in the preceding sections, we next present a carefully designed empirical study that enables a didactic comparison of the different deep unfolding paradigms. 
While in Section~\ref{sec:Unfolding} we illustrated each methodology through diverse examples drawn from various domains, here we focus on a single representative inference task based on the \ac{rpca} optimization problem. 
This controlled experimental setup allows us to directly compare the effect of the unfolding design choices under identical conditions, using the same base iterative optimizer and dataset\footnote{The quantitative study is available online at \url{https://gist.github.com/nirshlezinger1/212fedfb5ab363a507185935dd7c520d}}.  

\paragraph*{RPCA Optimization Problem}
The \ac{rpca} task aims to decompose a given data matrix $\RPCAMat \in \mathbb{R}^{n_1 \times n_2}$ into a low-rank matrix $\RPCARank$ and a component that is sparse in a known domain $\myMat{\Psi}$. 
Formally, the model assumes that
\begin{equation}
\RPCAMat = \RPCARank^\star + \myMat{\Psi}\RPCASparse^\star,
\label{eqn:RPCAModel}
\end{equation}
where $\RPCARank^\star$ is a rank $r$ matrix and $\RPCASparse^\star$ is $\kappa$-sparse, with $r,\kappa \ll \min(n_1,n_2)$. 
The objective of \ac{rpca} is to recover both $\RPCARank^\star$ and $\RPCASparse^\star$ from $\RPCAMat$, i.e., to determine the decision  $\Label = \{\RPCARank, \RPCASparse\}$ from the observed $\Input = \RPCAMat$. 

Several optimization formulations have been proposed for tackling the \ac{rpca} task (see, e.g.,~\cite{candes2011robust}). 
Here, we adopt one of~\cite{yi2016fast}, which factorizes the low-rank matrix as $\RPCARank = \myMat{L}\myMat{R}^\top$, where $\myMat{L} \in \mathbb{R}^{n_1 \times r}$ and $\myMat{R} \in \mathbb{R}^{n_2 \times r}$. 
The resulting optimization problem is given by
\begin{align}
\label{eqn:RPCA_nonconvex}
&\min_{\myMat{L}, \myMat{R}, \RPCASparse} 
\mathcal{L}_{\ObjParam}(\myMat{L},\myMat{R},\RPCASparse; \RPCAMat) 
\triangleq 
\frac{1}{2}\|\myMat{L}\myMat{R}^\top + \myMat{\Psi}\RPCASparse - \RPCAMat\|_F^2, 
\quad \text{s.t. } \|\RPCASparse\|_0 \leq \kappa.
\end{align}
The objective parameters $\ObjParam$ of 
\eqref{eqn:RPCA_nonconvex} include the transformation $\myMat{\Psi}$, as well as the rank and sparsity limits $r,\kappa$.

\paragraph*{Iterative Optimizer}
The optimization problem~\eqref{eqn:RPCA_nonconvex} can be addressed using an iterative algorithm that relaxes the $\ell_0$ constraint in \eqref{eqn:RPCA_nonconvex}, and alternates between updates of the sparse component $\RPCASparse$ and the low-rank factors $\myMat{L}$ and $\myMat{R}$~\cite{yi2016fast,cai2021learned}. 
That is,  we consider the 
$\ell_{1}$-regularized relaxation
\begin{equation}
    \min_{\myMat{L}, \myMat{R}, \RPCASparse} \;
       \frac{1}{2}\|\myMat{L}\myMat{R}^\top + \myMat{\Psi}\RPCASparse - \RPCAMat\|_F^2
        + \lambda \|\RPCASparse\|_{1},
    \label{eq:RPCA_relaxed}
\end{equation}
where $\lambda>0$ controls the sparsity level. Under this relaxation, the update of $\RPCASparse$ becomes a proximal step for the $\ell_{1}$ term, yielding the soft-thresholding rule. Thus, 
under the assumption that $\myMat{\Psi}$ is an orthonormal transform (i.e., $\myMat{\Psi}^\top \myMat{\Psi} = \mathbf{I}$),
at iteration~$k$, the sparse component is updated as
\begin{equation}
\label{eq:outlier_update}
\RPCASparse^{(k+1)} = 
\mySet{T}_{\zeta}\left(\myMat{\Psi}^{-1}\left(
\RPCAMat - \myMat{L}^{(k)} (\myMat{R}^{(k)})^\top\right)
\right),
\end{equation}
where $\mySet{T}_{\zeta}(\cdot)$ is the element-wise soft-thresholding operator with threshold hyperparameter $\zeta$. 
The low-rank factors are then updated using scaled gradient steps:
\begin{align}
\myMat{L}^{(k+1)}  
&= \myMat{L}^{(k)} 
- \eta_L
\cdot
\nabla_{\myMat{L}} \mathcal{L}_{\ObjParam}(\myMat{L}^{(k)}, \myMat{R}^{(k)}, \RPCASparse^{(k+1)}; \RPCAMat)
\left((\myMat{R}^{(k)})^\top \myMat{R}^{(k)}\right)^{-1}
, 
\label{eq:update_L}
\\
\myMat{R}^{(k+1)} 
&= \myMat{R}^{(k)} 
- \eta_R 
\cdot
\nabla_{\myMat{R}} \mathcal{L}_{\ObjParam}(\myMat{L}^{(k)}, \myMat{R}^{(k)}, \RPCASparse^{(k+1)}; \RPCAMat)
\left((\myMat{L}^{(k+1)})^\top \myMat{L}^{(k+1)}\right)^{-1}, 
\label{eq:update_R}
\end{align}
where $\eta_L$ and $\eta_R$ are scalar step-size hyperparameters. 
The solver hyperparameters $\HypParam$ thus include the step-sizes $\{\eta_L, \eta_R\}$, as well as the soft threshold $\zeta$ (which should conceptually be tuned based on the sparsity objective parameters $\kappa$, and thus can be viewed as either an implicit objective parameter or a hyperparameter of the solver, as we do here).
This iterative procedure forms the basis for all unfolded architectures examined in the sequel, each differing only in how the iteration-specific parameters are represented and learned.

\paragraph*{Unfolded Optimizers}
To enable a systematic comparison among the deep unfolding methodologies, we construct four unfolded variants of the iterative \ac{rpca} optimizer described above, each corresponding to one of the methodologies detailed in Section~\ref{sec:Unfolding}. 
All unfolded architectures are implemented with  $K=10$ iterations, and differ solely in the nature of their trainable parameters and the form of their iterative mappings.

\begin{enumerate}
    \item {\bf Learning Hyperparameters.} The first variant   unfolds the iterative \ac{rpca} solver with iteration-specific step sizes and thresholds. 
This approach follows the  design proposed in~\cite{cai2021learned}, in which the hyperparameters of each iteration $k$, denoted 
$\HypParam_k = \{\eta_L^{(k)}, \eta_R^{(k)}, \zeta^{(k)}\}$,
are treated as trainable parameters.  
\item {\bf Learning Objective Parameters.}
In the second variant, 
the transform $\myMat{\Psi}$ 
in~\eqref{eqn:RPCA_nonconvex} is allowed to vary across iterations and is learned from data. 
Accordingly, the parameters of the $k$th iteration are 
$\dnnParam_k = \{\myMat{\Psi}_k, \HypParam_k\}$.
%
\item {\bf Learning Correction Term.}
The third approach augments the iterative optimizer by learning data-driven corrections to its update rules. 
Here, the updates of $\myMat{L}$ and $\myMat{R}$ in~\eqref{eq:update_L}-\eqref{eq:update_R} are complemented with additive correction terms produced by an external neural network, in addition to iteration-specific hyperparameters. 
Specifically, we employ a {two-layer convolutional neural network} $\Delta(\cdot ; \dnnParam)$ for each iteration $k$, that takes as input the current estimates $\myMat{L}^{(k)}, \myMat{R}^{(k)}, \RPCASparse^{(k)}$  and the observation $\RPCAMat$, and outputs correction tensors that are added to the computed gradient steps. 
\item {\bf \ac{dnn} Inductive Bias.}
The final methodology 
replaces gradient-based updates with learned neural mappings. The resulting architecture is based on the one used for learning objective parameters, while having the transformation $\myMat{\Psi}$ be implemented using a two layer convolutional neural network that takes as input the current estimates $\myMat{L}^{(k)}, \myMat{R}^{(k)}, \RPCASparse^{(k)}$  and the observation $\RPCAMat$. 
\end{enumerate}

 \textcolor{NewColor}{These methodologies span a wide range of parameterizations: while learning only hyperparameters results in highly parameter-efficient models with only $30$ trainable parameters, incorporating learned correction terms ($3 \cdot 10^6$ parameters) or \ac{dnn}-based inductive bias ($10^7$ parameters)  leads to significantly larger models, reflecting the increased flexibility and abstraction of these approaches.}

\paragraph*{Numerical Evaluation}
We use the simulation scenario detailed in \cite{cai2021learned},  in which the data matrix sizes are $n_1=n_2=1000$, the low-rank component has $r=5$, and $10\%$ of the entries of the sparse component are non-zero.  
\textcolor{NewColor}{We consider two numerical studies: comparing different unfolded architectures arising from the same optimizer (under full and mismatched knowledge of $\mySet{L}_{\ObjParam}$), and comparing different training losses of \eqref{eqn:LossTypes} for a single architecture. In both cases, all architectures are  trained using the same dataset of RPCA observations $\{\RPCAMat_i\}$, and are optimized via a supervised loss \eqref{eqn:SupervisedLoss} based on the Frobenius norm deviation in the low rank component, enabling systematic comparison in controlled settings.}

\textcolor{NewColor}{We first compare the unfolded architectures.} 
Training is carried out in two stages, corresponding to the schemes outlined in Section~\ref{sec:Unfolding}. 
First, {\em sequential training} (\ref{itm:sequential}) is applied, where each iteration  is trained independently to minimize the objective over its corresponding output while the preceding iterations are fixed.  
Subsequently, the entire unfolded model is fine-tuned through {\em end-to-end training} (\ref{itm:e2e}), allowing global adjustment of all iteration-specific parameters based on the final reconstruction loss.  For the model-based optimizer, the hyperparameters $\HypParam$ are tuned manually to optimize the loss achieved with sufficient iterations (which here required over $10^5$ iterations to converge). 

The resulting loss values, \textcolor{NewColor}{i.e., the  average Frobenius reconstruction error of the low-rank component}, versus iteration of the unfolded optimizers compared to the original model-based optimizer are reported in Fig.~\ref{fig:RPCA_Res}, \textcolor{NewColor}{depicting a zoomed-in view of the initial $K=10$ iterations.}. Specifically, in Fig.~\ref{fig:RPCA_NoMimatch_1} we see that when provided with an accurate description of the optimization (i.e., the same $\ObjParam$ as the one used for generating the data), all unfolded optimizers achieve roughly the same performance, and their gain over the model-based optimizer is in their substantial improvement in latency. It is noted though that \ac{dnn}-aided methodologies of learned correction term and \ac{dnn} inductive bias both involve higher parameterization and more complex training. However, when the algorithms are provided with 
a noisy version of $\ObjParam$, we observe in Fig.~\ref{fig:RPCA_Mismatch} that different unfolding methods lead to different performance levels. Specifically, 
the architectures that dos not rely on explicit parameterization of the objective, i.e., learned objective, notably outperforms. It is also noted that the \ac{dnn} inductive bias model, which only uses $\ObjParam$ in its initialization, struggles in learning to deviate substantially from the mismatched objective, and is outperformed by learned objective. Still, all unfolded models notably outperform the model-based optimizer which struggles in terms of both latency and in dealing with mismatched objective. 


\begin{figure}
\begin{subfigure}{0.48\linewidth}
        \centering
    \includegraphics[width=\linewidth]{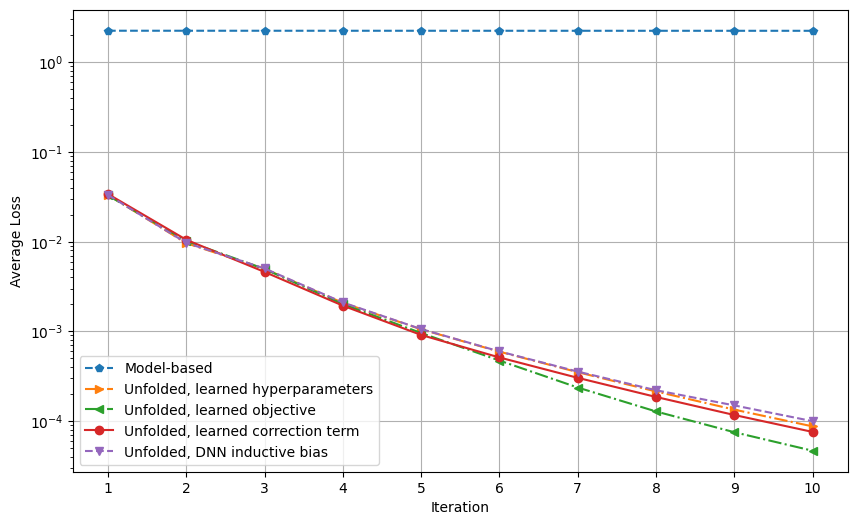}
    \caption{Full knowledge of $\mySet{L}_{\ObjParam}$}
    \label{fig:RPCA_NoMimatch_1}
\end{subfigure}
\hfill
\begin{subfigure}{0.48\linewidth}
        \centering
    \includegraphics[width=\linewidth]{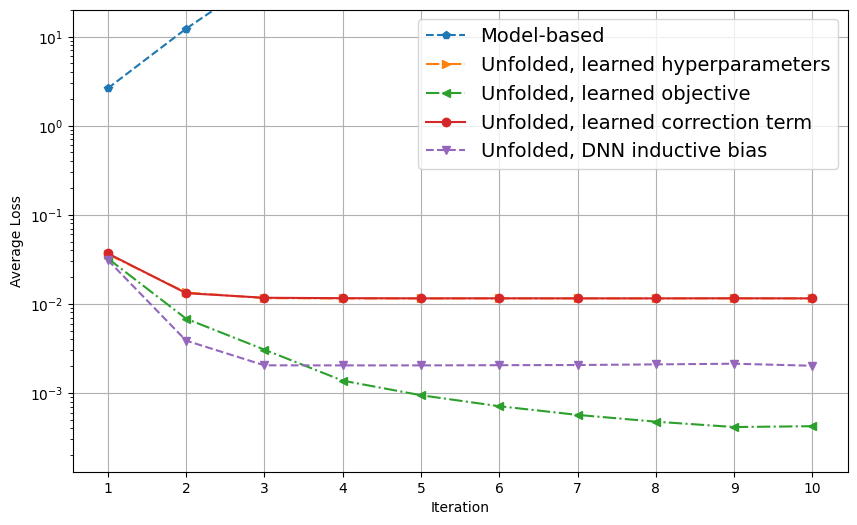}
    \caption{Mismatched $\mySet{L}_{\ObjParam}$} 
    \label{fig:RPCA_Mismatch}
\end{subfigure}
    \caption{Loss vs. iteration for RPCA setup of different unfolded optimizers}
    \label{fig:RPCA_Res}
\end{figure}

\begin{wrapfigure}{r}{0.48\linewidth} 
    \centering
    \vspace{-0.4cm}
    \includegraphics[width=\linewidth]{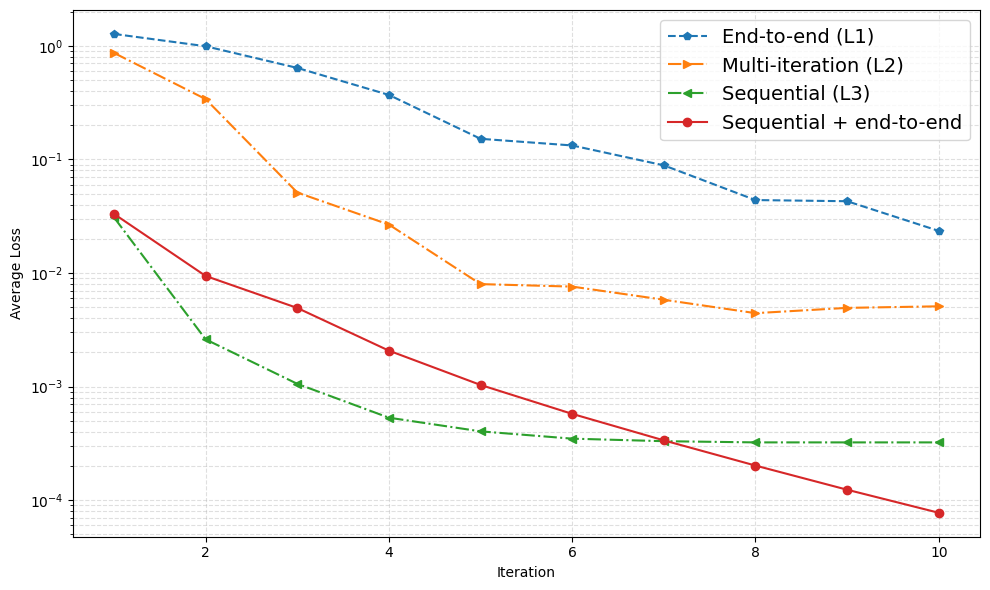}
    \caption{\textcolor{NewColor}{Training losses comparison for learned hyperparameters unfolded optimizers,  RPCA setup}}
    \label{fig:training_type_survey}
    \vspace{-0.4cm}
\end{wrapfigure}
\textcolor{NewColor}{Next, we compare the different individual training strategies \ref{itm:e2e}-\ref{itm:sequential}, as well the combined (\ref{itm:sequential} followed by \ref{itm:e2e} as used in the previous study) for tuning an unfolded optimizer with learned hyperparameters with fully known  $\mySet{L}_{\ObjParam}$. The results, shown  in Fig.~\ref{fig:training_type_survey}, highlight the individual contribution of each training component. Specifically, end-to-end training (\ref{itm:e2e}) by itself struggles to achieve high performance due to the difficulty of jointly optimizing the parameters. This limitation is partially alleviated by the multi-iteration loss (\ref{itm:multiiter}), which provides more direct supervision to the intermediate layers and thereby improves the learning dynamics. Sequential training (\ref{itm:sequential}) enables a more gradual and balanced learning process, as each iteration is trained in a structured manner before moving to later stages. Still, treating each iteration separately may limit the final achievable performance. Finally, combining sequential training with end-to-end fine-tuning yields the best overall results, since it benefits both from the stability of progressive iterative learning and from the global refinement enabled by joint optimization. 
}

\section{Future Research Directions}
\label{sec:future_research}
 
The tutorial-style overview and comparative study presented in this article highlight that deep unfolding is far more than a mere architectural design choice, but  a unifying framework for bridging model-based optimization and data-driven learning. 
The systematic categorization, theoretical insights, and empirical results discussed herein also reveal several open challenges and promising avenues for future exploration. 
Advancing the field of deep unfolding thus requires a deeper understanding of its theoretical foundations, extensions to more complex problem domains, and the development of practical methodologies that translate its conceptual potential into real-world performance gains. 
Below, we outline several core research directions that can further unveil the strengths and prospective use cases of deep unfolded designs.

\paragraph{Theoretical Foundations and Generalization Analysis}
While deep unfolding has demonstrated remarkable empirical success, a comprehensive theoretical understanding of its properties beyond the existing findings discussed in Section~\ref{sec:Theoretical} remains an open challenge. 
Future research should aim to establish rigorous guarantees on convergence, stability, and generalization under realistic nonconvex and stochastic conditions. 
Particularly, there is a need to characterize how the introduction of learned parameters and data-driven adaptations affects traditional optimization properties such as monotonic descent, fixed-point behavior, and robustness to perturbations. 
Bridging these insights with modern learning theory could provide a formal basis for explaining the empirical reliability of unfolded architectures.

\paragraph{Surrogate Objectives and Learned Optimization Goals}
A key aspect of deep unfolding lies in the potential to learn per-iteration surrogate objectives that deviate from the original mathematical formulation. 
Understanding when and why such modifications improve performance, and how they relate to the true task objectives, remains a rich topic for exploration. 
Developing theoretical and empirical tools for analyzing learned surrogate functions could lead to new principles for balancing interpretability and abstraction, guiding the design of optimizers that adapt their objectives intelligently throughout the inference process.

\paragraph{Extensions to Distributed and Closed-Loop Systems}
Deep unfolding is particularly promising for complex environments where optimization and learning are intertwined across multiple interacting entities. 
Extending unfolding frameworks to {\em distributed systems}~\cite{boyd2011distributed}, where local agents cooperatively optimize shared objectives under communication constraints, has only been explored in very specific settings to date, e.g., \cite{noah2024distributed}. 
Similarly, integrating deep unfolding within {\em closed-loop systems}~\cite{agrawal2020learning}, such as adaptive control and feedback-based signal processing, opens new opportunities for data-driven design of optimization-based control policies that remain interpretable and verifiable.

\paragraph{Application-Specific and Domain-Aware Design}
Another important research direction lies in aligning deep unfolding architectures with domain-specific constraints and goals. 
Many practical applications present unique physical and statistical structures that can be exploited to guide the unfolding process. 
Developing principled design methodologies that adapt unfolding strategies to the characteristics of each application will be crucial for achieving robust, interpretable, and efficient operation in realistic environments.

\paragraph{Hardware-Aware and Energy-Efficient Unfolding}
As deep unfolded models move closer to deployment, there is a growing need for {\em hardware-aware unfolding}, where the learning process is designed with computational and energy constraints in mind. 
Future research should explore techniques that  yield optimizers that are not only theoretically sound but also inherently compliant with device limitations. 
In the longer term, unfolding principles could even inform the {\em co-design of algorithms and hardware}, inspiring specialized architectures that exploit the structured and iterative nature of deep unfolded computation.

\paragraph{Robust and Continual Learning in Dynamic Environments}
Finally, extending deep unfolding to handle dynamic or non-stationary environments is a  vital research direction. 
Incorporating ideas from continual learning, online adaptation, and robust training could yield unfolded optimizers that adapt seamlessly to changing data distributions and system conditions. 
Such advances would be  impactful in resource-constrained and mission-critical applications, where the ability to learn and adjust on-the-fly is essential.




\balance
\bibliographystyle{IEEEtran}
\bibliography{IEEEabrv,refs}

\begin{thebibliography}{10}
\providecommand{\url}[1]{#1}
\csname url@samestyle\endcsname
\providecommand{\newblock}{\relax}
\providecommand{\bibinfo}[2]{#2}
\providecommand{\BIBentrySTDinterwordspacing}{\spaceskip=0pt\relax}
\providecommand{\BIBentryALTinterwordstretchfactor}{4}
\providecommand{\BIBentryALTinterwordspacing}{\spaceskip=\fontdimen2\font plus
\BIBentryALTinterwordstretchfactor\fontdimen3\font minus \fontdimen4\font\relax}
\providecommand{\BIBforeignlanguage}[2]{{%
\expandafter\ifx\csname l@#1\endcsname\relax
\typeout{** WARNING: IEEEtran.bst: No hyphenation pattern has been}%
\typeout{** loaded for the language `#1'. Using the pattern for}%
\typeout{** the default language instead.}%
\else
\language=\csname l@#1\endcsname
\fi
#2}}
\providecommand{\BIBdecl}{\relax}
\BIBdecl

\bibitem{luo2006introduction}
Z.-Q. Luo and W.~Yu, ``An introduction to convex optimization for communications and signal processing,'' \emph{{IEEE} J. Sel. Areas Commun.}, vol.~24, no.~8, pp. 1426--1438, 2006.

\bibitem{boyd2004convex}
S.~P. Boyd and L.~Vandenberghe, \emph{Convex Optimization}.\hskip 1em plus 0.5em minus 0.4em\relax Cambridge University Press, 2004.

\bibitem{shlezinger2022model}
N.~Shlezinger, Y.~C. Eldar, and S.~P. Boyd, ``Model-based deep learning: On the intersection of deep learning and optimization,'' \emph{{IEEE} Access}, vol.~10, pp. 115\,384--115\,398, 2022.

\bibitem{dahrouj2021overview}
H.~Dahrouj, R.~Alghamdi, H.~Alwazani, S.~Bahanshal, A.~A. Ahmad, A.~Faisal, R.~Shalabi, R.~Alhadrami, A.~Subasi, M.~T. Al-Nory, O.~Kittaneh, and J.~S. Shamma, ``An overview of machine learning-based techniques for solving optimization problems in communications and signal processing,'' \emph{{IEEE} Access}, vol.~9, pp. 74\,908--74\,938, 2021.

\bibitem{zappone2019wireless}
A.~Zappone, M.~Di~Renzo, and M.~Debbah, ``Wireless networks design in the era of deep learning: Model-based, {AI}-based, or both?'' \emph{{IEEE} Trans. Commun.}, vol.~67, no.~10, pp. 7331--7376, 2019.

\bibitem{hershey2014deep}
J.~R. Hershey, J.~L. Roux, and F.~Weninger, ``Deep unfolding: Model-based inspiration of novel deep architectures,'' \emph{arXiv preprint arXiv:1409.2574}, 2014.

\bibitem{monga2021algorithm}
V.~Monga, Y.~Li, and Y.~C. Eldar, ``Algorithm unrolling: Interpretable, efficient deep learning for signal and image processing,'' \emph{{IEEE} Signal Process. Mag.}, vol.~38, no.~2, pp. 18--44, 2021.

\bibitem{gregor2010learning}
K.~Gregor and Y.~LeCun, ``Learning fast approximations of sparse coding,'' in \emph{International Conference on Machine Learning}, 2010, pp. 399--406.

\bibitem{krizhevsky2012imagenet}
A.~Krizhevsky, I.~Sutskever, and G.~E. Hinton, ``Imagenet classification with deep convolutional neural networks,'' \emph{Advances in Neural Information Processing Systems}, vol.~25, 2012.

\bibitem{shlezinger2023model}
N.~Shlezinger, J.~Whang, Y.~C. Eldar, and A.~G. Dimakis, ``Model-based deep learning,'' \emph{Proc. {IEEE}}, vol. 111, no.~5, pp. 465--499, 2023.

\bibitem{khani2020adaptive}
M.~Khani, M.~Alizadeh, J.~Hoydis, and P.~Fleming, ``Adaptive neural signal detection for massive {MIMO},'' \emph{{IEEE} Trans. Wireless Commun.}, vol.~19, no.~8, pp. 5635--5648, 2020.

\bibitem{chowdhury2021unfolding}
A.~Chowdhury, G.~Verma, C.~Rao, A.~Swami, and S.~Segarra, ``Unfolding {WMMSE} using graph neural networks for efficient power allocation,'' \emph{{IEEE} Trans. Wireless Commun.}, vol.~20, no.~9, pp. 6004--6017, 2021.

\bibitem{samuel2019learning}
N.~Samuel, T.~Diskin, and A.~Wiesel, ``Learning to detect,'' \emph{{IEEE} Trans. Signal Process.}, vol.~67, no.~10, pp. 2554--2564, 2019.

\bibitem{shlezinger2020deepsic}
N.~Shlezinger, R.~Fu, and Y.~C. Eldar, ``Deep{SIC}: Deep soft interference cancellation for multiuser {MIMO} detection,'' \emph{{IEEE} Trans. Wireless Commun.}, vol.~20, no.~2, pp. 1349--1362, 2020.

\bibitem{lavi2023learn}
O.~Lavi and N.~Shlezinger, ``Learn to rapidly and robustly optimize hybrid precoding,'' \emph{{IEEE} Trans. Commun.}, vol.~71, no.~10, pp. 5814--5830, 2023.

\bibitem{noah2024distributed}
Y.~Noah and N.~Shlezinger, ``Distributed learn-to-optimize: Limited communications optimization over networks via deep unfolded distributed {ADMM},'' \emph{{IEEE} Trans. Mobile Comput.}, vol.~24, no.~2, pp. 3012--3024, 2025.

\bibitem{raviv2023online}
T.~Raviv, S.~Park, O.~Simeone, Y.~C. Eldar, and N.~Shlezinger, ``Online meta-learning for hybrid model-based deep receivers,'' \emph{{IEEE} Trans. Wireless Commun.}, vol.~22, no.~10, pp. 6415--6431, 2023.

\bibitem{saravanos2024deep}
A.~D. Saravanos, H.~Kuperman, A.~Oshin, A.~T. Abdul, V.~Pacelli, and E.~Theodorou, ``Deep distributed optimization for large-scale quadratic programming,'' in \emph{International Conference on Learning Representations (ICLR)}, 2025.

\bibitem{avrahami2025deep}
D.~Avrahami, A.~Milstein, C.~Chaux, T.~Routtenberg, and N.~Shlezinger, ``Deep unfolding with approximated computations for rapid optimization,'' \emph{arXiv preprint arXiv:2509.00782}, 2025.

\bibitem{sofer2025unveiling}
E.~Sofer, T.~Shaked, C.~Chaux, and N.~Shlezinger, ``Unveiling and mitigating adversarial vulnerabilities in iterative optimizers,'' \emph{{IEEE} Trans. Signal Process.}, 2025.

\bibitem{hadou2024robust}
S.~Hadou, N.~NaderiAlizadeh, and A.~Ribeiro, ``Robust stochastically-descending unrolled networks,'' \emph{{IEEE} Trans. Signal Process.}, vol.~72, pp. 5484--5499, 2024.

\bibitem{shah2024optimization}
S.~B. Shah, P.~Pradhan, W.~Pu, R.~Randhi, M.~R. Rodrigues, and Y.~C. Eldar, ``Optimization guarantees of unfolded {ISTA} and {ADMM} networks with smooth soft-thresholding,'' \emph{{IEEE} Trans. Signal Process.}, vol.~72, pp. 3272--3286, 2024.

\bibitem{scarlett2022theoretical}
J.~Scarlett, R.~Heckel, M.~R.~D. Rodrigues, P.~Hand, and Y.~C. Eldar, ``Theoretical perspectives on deep learning methods in inverse problems,'' \emph{IEEE Journal on Selected Areas in Information Theory}, vol.~3, no.~3, pp. 433--453, 2022.

\bibitem{nareddy2025some}
K.~K.~R. Nareddy, I.~Perumal, and C.~S. Seelamantula, ``Some intriguing observations on the learnt matrices in deep unfolded networks,'' in \emph{IEEE International Conference on Acoustics, Speech and Signal Processing (ICASSP)}, 2025.

\bibitem{candes2011robust}
E.~J. Cand{\`e}s, X.~Li, Y.~Ma, and J.~Wright, ``Robust principal component analysis?'' \emph{Journal of the ACM}, vol.~58, no.~3, pp. 1--37, 2011.

\bibitem{Luenberger2008}
D.~G. Luenberger and Y.~Ye, \emph{Linear and Nonlinear Programming}, 3rd~ed.\hskip 1em plus 0.5em minus 0.4em\relax New York, NY: Springer, 2008.

\bibitem{Kayestimation}
S.~M. Kay, \emph{Fundamentals of statistical signal processing: Estimation Theory}.\hskip 1em plus 0.5em minus 0.4em\relax Englewood Cliffs (N.J.): Prentice Hall PTR, 1993, vol.~1.

\bibitem{EldarKutyniok2012}
Y.~C. Eldar and G.~Kutyniok, Eds., \emph{Compressed Sensing: Theory and Applications}.\hskip 1em plus 0.5em minus 0.4em\relax Cambridge University Press, 2012.

\bibitem{chen2022learning}
T.~Chen, X.~Chen, W.~Chen, H.~Heaton, J.~Liu, Z.~Wang, and W.~Yin, ``Learning to optimize: A primer and a benchmark,'' \emph{Journal of Machine Learning Research}, vol.~23, no. 189, pp. 1--59, 2022.

\bibitem{song2024towards}
Q.~Song, W.~Lin, J.~Wang, and H.~Xu, ``Towards robust learning to optimize with theoretical guarantees,'' in \emph{Proceedings of the IEEE/CVF Conference on Computer Vision and Pattern Recognition}, 2024, pp. 27\,498--27\,506.

\bibitem{bai2019deep}
S.~Bai, J.~Z. Kolter, and V.~Koltun, ``Deep equilibrium models,'' \emph{Advances in Neural Information Processing Systems}, 2019.

\bibitem{ahmad2020plug}
R.~Ahmad, C.~A. Bouman, G.~T. Buzzard, S.~Chan, S.~Liu, E.~T. Reehorst, and P.~Schniter, ``Plug-and-play methods for magnetic resonance imaging: Using denoisers for image recovery,'' \emph{{IEEE} Signal Process. Mag.}, vol.~37, no.~1, pp. 105--116, 2020.

\bibitem{zhang2025feasibility}
H.~Zhang and H.~Sun, ``Feasibility guaranteed learning-to-optimize in wireless communication resource allocation,'' \emph{arXiv preprint arXiv:2509.02417}, 2025.

\bibitem{hadou2025unrolled}
S.~Hadou and A.~Ribeiro, ``Unrolled graph neural networks for constrained optimization,'' \emph{arXiv preprint arXiv:2509.17156}, 2025.

\bibitem{nocedal1999numerical}
J.~Nocedal and S.~J. Wright, \emph{Numerical optimization}.\hskip 1em plus 0.5em minus 0.4em\relax Springer, 1999.

\bibitem{cavalcanti2025adaptive}
J.~V. Cavalcanti, L.~Lessard, and A.~C. Wilson, ``Adaptive backtracking line search,'' in \emph{International Conference on Learning Representations (ICLR)}, 2025.

\bibitem{khobahi2021lord}
S.~Khobahi, N.~Shlezinger, M.~Soltanalian, and Y.~C. Eldar, ``Lo{RD-N}et: Unfolded deep detection network with low-resolution receivers,'' \emph{{IEEE} Trans. Signal Process.}, vol.~69, pp. 5651--5664, 2021.

\bibitem{yu2023white}
Y.~Yu, S.~Buchanan, D.~Pai, T.~Chu, Z.~Wu, S.~Tong, B.~Haeffele, and Y.~Ma, ``White-box transformers via sparse rate reduction,'' \emph{Advances in Neural Information Processing Systems}, vol.~36, pp. 9422--9457, 2023.

\bibitem{do2024interpretable}
T.~T. Do, P.~Eftekhar, S.~A. Hosseini, G.~Cheung, and P.~A. Chou, ``Interpretable lightweight transformer via unrolling of learned graph smoothness priors,'' \emph{Advances in Neural Information Processing Systems}, vol.~37, pp. 6393--6416, 2024.

\bibitem{chowdhury2023deep}
A.~Chowdhury, G.~Verma, A.~Swami, and S.~Segarra, ``Deep graph unfolding for beamforming in {MU-MIMO} interference networks,'' \emph{{IEEE} Trans. Wireless Commun.}, vol.~23, no.~5, pp. 4889--4903, 2023.

\bibitem{choi2000}
W.-J. Choi, K.-W. Cheong, and J.~Cioffi, ``Iterative soft interference cancellation for multiple antenna systems,'' in \emph{Proc. IEEE WCNC}, 2000.

\bibitem{revach2023rtsnet}
G.~Revach, X.~Ni, N.~Shlezinger, R.~J. Van~Sloun, and Y.~C. Eldar, ``{RTSN}et: Learning to smooth in partially known state-space models,'' \emph{{IEEE} Trans. Signal Process.}, vol.~71, pp. 4441--4456, 2023.

\bibitem{gusakov2025rapid}
Y.~Gusakov, O.~Simeone, T.~Routtenberg, and N.~Shlezinger, ``Rapid online {B}ayesian learning for deep receivers,'' in \emph{Proc. IEEE ICASSP}, 2025.

\bibitem{uzlaner2025async}
N.~Uzlaner, T.~Raviv, N.~Shlezinger, and K.~Todros, ``Asynchronous online adaptation via modular drift detection for deep receivers,'' \emph{{IEEE} Trans. Wireless Commun.}, vol.~24, no.~5, pp. 4454--4468, 2025.

\bibitem{raviv2025modular}
T.~Raviv and N.~Shlezinger, ``Modular hypernetworks for scalable and adaptive deep {MIMO} receivers,'' \emph{IEEE Open Journal of Signal Processing}, vol.~6, pp. 256--265, 2025.

\bibitem{agrawal2021learning}
A.~Agrawal, S.~Barratt, and S.~Boyd, ``Learning convex optimization models,'' \emph{{IEEE/CAA} J. Autom. Sinica}, vol.~8, no.~8, pp. 1355--1364, 2021.

\bibitem{craven1986nondifferentiable}
B.~D. Craven, ``Nondifferentiable optimization by smooth approximations,'' \emph{Optimization}, vol.~17, no.~1, pp. 3--17, 1986.

\bibitem{janjuvsevic2026self}
N.~Janju{\v{s}}evi{\'c}, J.~Chen, L.~Ginocchio, M.~Bruno, Y.~Huang, Y.~Wang, H.~Chandarana, and L.~Feng, ``Self-supervised noise adaptive {MRI} denoising via repetition to repetition ({Rep2Rep}) learning,'' \emph{Magnetic Resonance in Medicine}, vol.~95, no.~3, pp. 1619--1633, 2026.

\bibitem{chen2018theoretical}
X.~Chen, J.~Liu, Z.~Wang, and W.~Yin, ``Theoretical linear convergence of unfolded {ISTA} and its practical weights and thresholds,'' in \emph{Advances in {Neural} {Information} {Processing} {Systems}}, vol.~31.\hskip 1em plus 0.5em minus 0.4em\relax Curran Associates, Inc., 2018.

\bibitem{liu2019alista}
J.~Liu, X.~Chen, Z.~Wang, and W.~Yin, ``{ALISTA}: {Analytic} weights are as good as learned weights in {LISTA},'' in \emph{International {Conference} on {Learning} {Representations} (ICLR)}, 2019.

\bibitem{chen2021hyperparameter}
X.~Chen, J.~Liu, Z.~Wang, and W.~Yin, ``Hyperparameter tuning is all you need for {LISTA},'' \emph{Advances in Neural Information Processing Systems}, vol.~34, pp. 11\,678--11\,689, 2021.

\bibitem{yang2025deep}
J.~Yang, B.~Ai, W.~Chen, S.~Yang, N.~Wang, and C.~Yuen, ``Deep unfolding-based sensing-assisted channel estimation with imperfect radar arrays,'' \emph{{IEEE} J. Sel. Areas Commun.}, 2025.

\bibitem{yang2025improving}
L.~Yang, J.~Mi, W.~Li, G.~Wang, and B.~Xiao, ``Improving the sparse coding model via hybrid {Gaussian} priors,'' \emph{Pattern Recognition}, vol. 159, p. 111102, 2025.

\bibitem{heaton2023safeguarded}
H.~Heaton, X.~Chen, Z.~Wang, and W.~Yin, ``Safeguarded learned convex optimization,'' \emph{AAAI Conference on Artificial Intelligence}, vol.~37, no.~6, pp. 7848--7855, 2023.

\bibitem{behboodi2022compressive}
A.~Behboodi, H.~Rauhut, and E.~Schnoor, ``Compressive sensing and neural networks from a statistical learning perspective,'' in \emph{Compressed {Sensing} in {Information} {Processing}}.\hskip 1em plus 0.5em minus 0.4em\relax Cham: Springer International Publishing, 2022, pp. 247--277.

\bibitem{schnoor2023generalization}
E.~Schnoor, A.~Behboodi, and H.~Rauhut, ``Generalization error bounds for iterative recovery algorithms unfolded as neural networks,'' \emph{Information and Inference: A Journal of the IMA}, vol.~12, no.~3, pp. 2267--2299, 2023.

\bibitem{shultzman2023generalization}
A.~Shultzman, E.~Azar, M.~R.~D. Rodrigues, and Y.~C. Eldar, ``Generalization and estimation error bounds for model-based neural networks,'' in \emph{{International} {Conference} on {Learning} {Representations} (ICLR)}, 2023.

\bibitem{shalev2014understanding}
S.~Shalev-Shwartz and S.~Ben-David, \emph{Understanding machine learning: From theory to algorithms}.\hskip 1em plus 0.5em minus 0.4em\relax Cambridge university press, 2014.

\bibitem{chen2020understanding}
X.~Chen, Y.~Zhang, C.~Reisinger, and L.~Song, ``Understanding deep architecture with reasoning layer,'' in \emph{Advances in {Neural} {Information} {Processing} {Systems}}, vol.~33.\hskip 1em plus 0.5em minus 0.4em\relax Curran Associates, Inc., 2020, pp. 1240--1252.

\bibitem{sucker2023pac-bayesian}
M.~Sucker and P.~Ochs, ``{PAC}-{Bayesian} learning of optimization algorithms,'' in \emph{{International} {Conference} on {Artificial} {Intelligence} and {Statistics}}, ser. Proceedings of {Machine} {Learning} {Research}, vol. 206.\hskip 1em plus 0.5em minus 0.4em\relax PMLR, 2023, pp. 8145--8164.

\bibitem{karan2024unrolled}
A.~Karan, K.~Shah, S.~Chen, and Y.~C. Eldar, ``Unrolled denoising networks provably learn to perform optimal bayesian inference,'' in \emph{The Thirty-eighth Annual Conference on Neural Information Processing Systems}, 2024.

\bibitem{montanari2024statistically}
A.~Montanari and Y.~Wu, ``Statistically optimal firstorder algorithms: a proof via orthogonalization,'' \emph{Information and Inference: A Journal of the IMA}, vol.~13, no.~4, p. iaae027, 2024.

\bibitem{vardi2023implicit}
G.~Vardi, ``On the implicit bias in deep-learning algorithms,'' \emph{Communications of the ACM}, vol.~66, no.~6, pp. 86--93, 2023.

\bibitem{garrigos2023handbook}
G.~Garrigos and R.~M. Gower, ``Handbook of convergence theorems for (stochastic) gradient methods,'' \emph{arXiv preprint arXiv:2301.11235}, 2023.

\bibitem{simon2019rethinking}
D.~Simon and M.~Elad, ``Rethinking the {CSC} model for natural images,'' \emph{Advances in Neural Information Processing Systems}, vol.~32, 2019.

\bibitem{janjuvsevic2022cdlnet}
N.~Janju{\v{s}}evi{\'c}, A.~Khalilian-Gourtani, and Y.~Wang, ``{CDLNet}: Noise-adaptive convolutional dictionary learning network for blind denoising and demosaicing,'' \emph{IEEE Open Journal of Signal Processing}, vol.~3, pp. 196--211, 2022.

\bibitem{yi2016fast}
X.~Yi, D.~Park, Y.~Chen, and C.~Caramanis, ``Fast algorithms for robust {PCA} via gradient descent,'' \emph{Advances in Neural Information Processing Systems}, vol.~29, 2016.

\bibitem{cai2021learned}
H.~Cai, J.~Liu, and W.~Yin, ``Learned robust {PCA}: A scalable deep unfolding approach for high-dimensional outlier detection,'' in \emph{Advances in {Neural} {Information} {Processing} {Systems}}, 2021.

\bibitem{boyd2011distributed}
S.~Boyd, N.~Parikh, E.~Chu, B.~Peleato, and J.~Eckstein, ``Distributed optimization and statistical learning via the alternating direction method of multipliers,'' \emph{Foundations and Trends{\textregistered} in Machine learning}, vol.~3, no.~1, pp. 1--122, 2011.

\bibitem{agrawal2020learning}
A.~Agrawal, S.~Barratt, S.~Boyd, and B.~Stellato, ``Learning convex optimization control policies,'' in \emph{Learning for Dynamics and Control}.\hskip 1em plus 0.5em minus 0.4em\relax PMLR, 2020, pp. 361--373.

\end{thebibliography}

\end{document}